\documentclass[10pt,twocolumn,letterpaper]{article}

\usepackage{iccv}
\usepackage{times}
\usepackage{epsfig}
\usepackage{graphicx}
\usepackage{amsmath}
\usepackage{amssymb}

\usepackage{bbm}
\usepackage{booktabs}
\usepackage{microtype}
\usepackage{pifont}
\usepackage{color}
\usepackage{algorithm} 
\usepackage{algpseudocode} 
\usepackage[caption=false]{subfig}
\usepackage[accsupp]{axessibility}

\usepackage{soul}

\usepackage[pagebackref=true,breaklinks=true,letterpaper=true,colorlinks,bookmarks=false]{hyperref}

\iccvfinalcopy 

\def\iccvPaperID{9650} 

\ificcvfinal\fi

\definecolor{highlander_blue}{RGB}{0,61,165}

\newcommand*{\Scale}[2][4]{\scalebox{#1}{$#2$}}%

\newcommand{\cmark}{\ding{51}}%
\newcommand{\xmark}{\ding{55}}%

\begin{document}

\title{SUMMIT: Source-Free Adaptation of Uni-Modal Models to Multi-Modal Targets}

\author{Cody Simons$^{1}$\quad Dripta S. Raychaudhuri$^{1,2,* }$\quad Sk Miraj Ahmed$^{1}$\quad Suya You$^{3}$\\ Konstantinos Karydis$^{1}$\quad Amit K. Roy-Chowdhury$^{1}$\\
$^{1}$University of California, Riverside\quad $^{2}$AWS AI Labs\quad $^{3}$DEVCOM Army Research Laboratory\\
{\tt\small \{csimo005@,drayc001@,sahme047@,kkarydis@ece.,amitrc@ece.\}ucr.edu}\quad {\tt\small suya.you.civ@army.mil}
}

\maketitle
\ificcvfinal\thispagestyle{empty}\fi

\newcommand\blfootnote[1]{%
  \begingroup
  \renewcommand\thefootnote{}\footnote{#1}%
  \addtocounter{footnote}{-1}%
  \endgroup
}
\blfootnote{* Currently at AWS AI Labs. Work done while the author was at UCR.}

\begin{figure*}[t]
    \centering
    \includegraphics[width=0.8\textwidth]{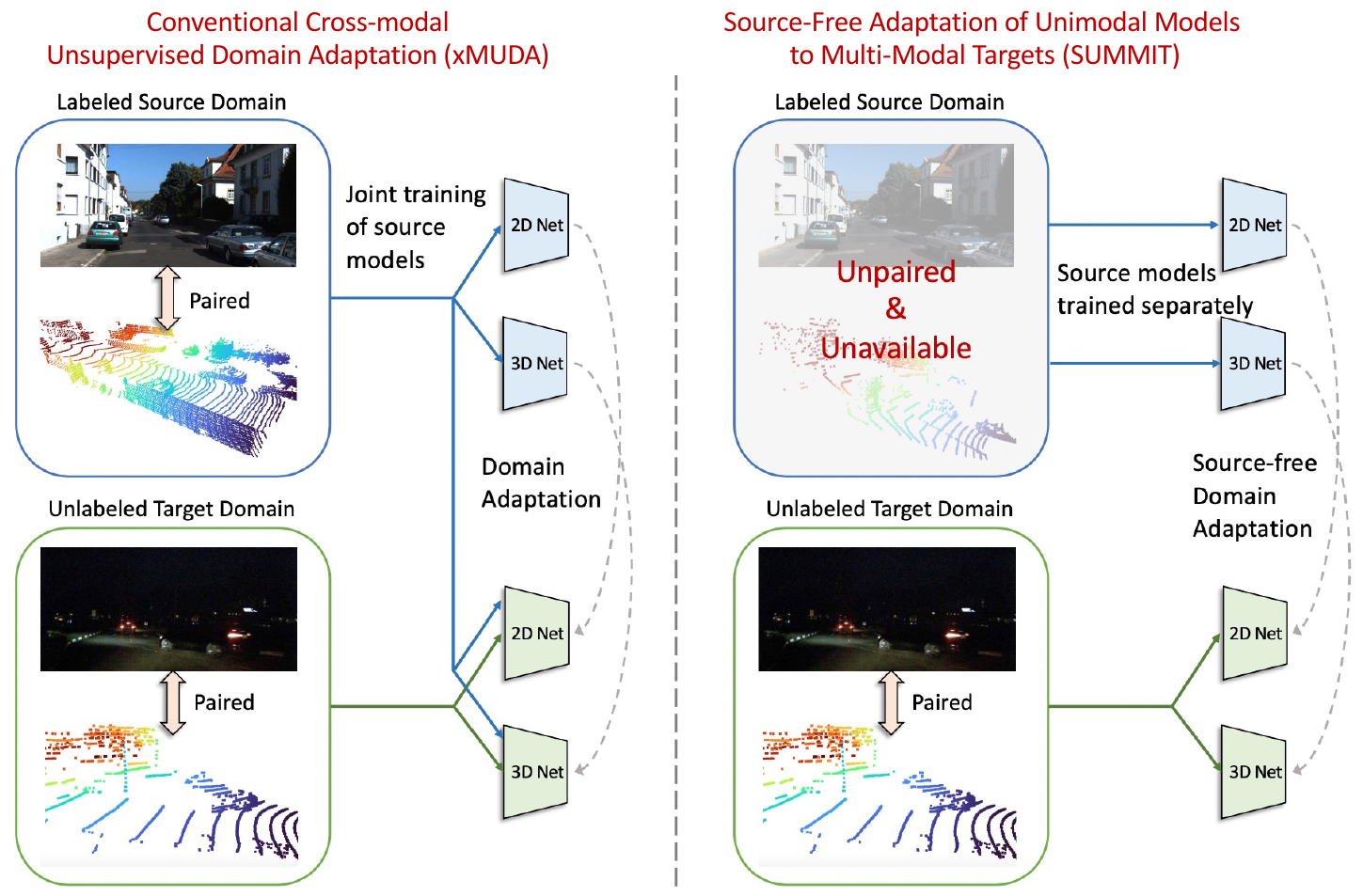}
    \caption{\textbf{Problem setup.} Our goal is to adapt a pair of uni-modal models, which have been trained independently on a source domain, to a target domain consisting of unlabeled, paired, multi-modal data, without access to the original source dataset. In contrast to conventional cross-modal UDA~\cite{jaritz2020xmuda} (left panel), we do not assume that the source dataset used for training (i) consists of paired multi-modal data, and (ii) is available during adaptation to the target domain. }
    \label{fig:teaser}
\end{figure*}

\begin{abstract}
Scene understanding using multi-modal data is necessary in many applications, e.g., autonomous navigation. To achieve this in a variety of situations, existing models must be able to adapt to shifting data distributions without arduous data annotation. Current approaches assume that the source data is available during adaptation and that the source consists of paired multi-modal data. Both these assumptions may be problematic for many applications. Source data may not be available due to privacy, security, or economic concerns. Assuming the existence of paired multi-modal data for training also entails significant data collection costs and fails to take advantage of widely available freely distributed pre-trained uni-modal models. In this work, we relax both of these assumptions by addressing the problem of adapting a set of models trained independently on uni-modal data to a target domain consisting of unlabeled multi-modal data, without having access to the original source dataset. Our proposed approach solves this problem through a switching framework which automatically chooses between two complementary methods of cross-modal pseudo-label fusion -- agreement filtering and entropy weighting -- based on the estimated domain gap. We demonstrate our work on the semantic segmentation problem. Experiments across seven challenging adaptation scenarios verify the efficacy of our approach, achieving results comparable to, and in some cases outperforming, methods which assume access to source data. Our method achieves an improvement in mIoU of up to 12\% over competing baselines. Our code is publicly available at \url{https://github.com/csimo005/SUMMIT}.
\end{abstract}

\section{Introduction}
There has been a recent surge of interest in autonomous vehicles which typically rely on a wide variety of sensors. This has fueled the need for machine learning models capable of processing multiple sensing modalities, commonly referred to as multi-modal models.  One problem of particular interest is 3D semantic segmentation, which has received a lot of interest \cite{graham20183d,qi2017pointnet++,wang2019graphcnn}\ driven by the introduction of new multi-modal datasets~\cite{caesar2020nuscenes,geyer2020a2d2,behley2019semantickitti}.
Many 3D semantic segmentation methods (e.g.,~\cite{graham20183d,qi2017pointnet++,wang2019graphcnn}) fuse data across different sensing modalities, e.g., RGB images and point clouds to obtain and employ colored pointclouds~\cite{teng2023centroid}, to increase performance and robustness. As with most learning problems, the performance of 3D semantic segmentation degrades as the input data distribution diverges from the training set distribution \cite{jaritz2020xmuda}, referred to as domain shift. This is particularly true for autonomous navigation which can experience domain shifts because of lighting and weather changes throughout the day, as well as geographic changes when traveling over large distances. Many works have sought to address this domain shift in the Unsupervised Domain Adaptation (UDA) setting~\cite{Hoffman_ICML_2018,mei2020instance,dong2020cscl}. Recently, the relationship between different modalities has been leveraged to aid in the adaptation process via cross-modal UDA (xMUDA)~\cite{jaritz2020xmuda}. 

While xMUDA~\cite{jaritz2020xmuda} has made significant improvements over uni-modal UDA methods, it assumes that the source dataset used for training (i) consists of paired multi-modal data, and (ii) is available during adaptation to the target domain.\footnote{In this work, modality refers to a specific type of input  data, such as RGB images or point clouds, while domain refers to the underlying data distribution, such as cities in different continents.} However, these conditions may be hard to satisfy in real-world scenarios: \\
$\bullet$ The first assumption is problematic because it requires collection and annotation of large volumes of paired multi-modal data for every sensor configuration (e.g., RGB and depth, RGB and IR, etc), a very time-consuming operation. Also, it fails to take advantage of the large volumes of uni-modal data and pre-trained models that are easily accessible.\\
$\bullet$ The second assumption is problematic because sharing the source data for adaptation may be impossible due to privacy, security and commercial reasons. Additionally, as datasets have grown, their transfer and storage have begun to present non-trivial engineering challenges and financial costs.

Relaxing the first assumption of paired modalities in the source domain remains unaddressed in cross-modal UDA. The second assumption has been relaxed in the source-free UDA setting~\cite{liang2020we,ahmed2021unsupervised,yeh2021sofa,xia2021adaptive}, but to the authors' best knowledge, no such prior works deal with multi-modal data. 

In this paper, we propose \textit{\underline{S}ource-free adaptation of \underline{U}ni-modal \underline{M}odels to \underline{M}ult\underline{i}-modal \underline{T}argets} (SUMMIT), relaxing both the above assumptions (see Figure~\ref{fig:teaser}). Relaxation of the two assumptions in conventional cross-modal UDA makes our problem substantially more challenging.
In the conventional cross-modal UDA setting, the approach in~\cite{jaritz2020xmuda} relies on labeled paired multi-modal data in the source domain to learn correlations across modalities. Learned correlations are then exploited to improve transfer to the target domain. In our setting, the correlations between modalities must be learned on the unlabeled target data, as we are working with uni-modal source models and do not assume access to such labeled pairs. The lack of source data already makes the alignment of the source and target distributions a challenging problem. Combining it with uni-modal models on the source side makes the overall problem even more challenging.

To address these challenges, we propose a new adaptation framework built on pseudo-label fusion across modalities. First, we utilize the trained uni-modal models to generate pseudo-labels on the target data, separately for each modality. Second, these pseudo-labels are fused together across modalities to filter out noisy predictions. We introduce a data-driven switching method that automatically chooses between two complementary approaches for cross-modal pseudo-label fusion -- \emph{agreement filtering} and \emph{entropy weighting} -- based on the estimated domain gap. The fused pseudo-labels are used to supervise the process of learning the correlations across modalities allowing for cross-modal learning to take place. Optimizing for explicit cross-modal objectives, our framework learns the correlations across modalities even without the presence of the source data and improves transfer beyond standard uni-modal adaptation.


\noindent \textbf{Main contributions.} Our primary contributions can be summarized as follows.\\
$\bullet$ We address the problem of adapting a set of models trained independently on uni-modal data to a target domain consisting of unlabeled, paired, multi-modal data, without access to the original source dataset. This setting of great practical importance as explained above. \\
$\bullet$ We propose a new cross-modal, source-free UDA framework which fuses pseudo-labels across modalities using information-theoretic and hypothesis testing approaches. This helps increase robustness of the predictions and implicitly allows for cross-modal correlations to be learned on the target domain without access to source data. \\
%
%
$\bullet$ We perform extensive experiments on seven challenging benchmarks which demonstrate that our method provides an improvement of up 12\% over competing baselines.
\section{Related Work} \label{sec:related}

\noindent \textbf{Unsupervised Domain Adaptation.} 
UDA methods have been applied to a wide variety of computer vision tasks, including image classification~\cite{tzeng2017adversarial}, semantic segmentation~\cite{tsai2018learning} and object detection~\cite{hsu2020progressive}, in an effort to address the data distribution shift. Most approaches try to align the source and target data distributions, using techniques such as maximum mean discrepancy~\cite{long2015learning} and adversarial learning~\cite{ganin2016domain,tzeng2017adversarial,raychaudhuri2021cross}. A separate line of work uses image translation techniques to perform adaptation by translating the source images to the target domain~\cite{Hoffman_ICML_2018,Luan_CVPR_2019}. In the case of semantic segmentation, existing UDA methods fall primarily into three categories: output space alignment, pixel-space alignment, and pseudo-labeling. The first category of methods aligns the output or feature distributions~\cite{tsai2018learning,araslanov2021self}. The pixel alignment approaches use image translation techniques similar to image classification~\cite{Hoffman_ICML_2018,Choi_ICCV19,Chang_CVPR_2019,yang2020fda}, while the pseudo-labeling techniques generate pixel-wise pseudo-labels to fine-tune the source model~\cite{Saleh_ECCV_2018,Zou_ECCV_2018,mei2020instance,dong2020cscl}. Some approaches also try to combine these strategies to perform adaptation~\cite{paul2020domain,musto2020semantically,subhani2020learning}. Our method is a UDA approach but in a multi-modal and source-free setting. 
\vskip 6pt
\noindent \textbf{Source-free Domain Adaptation.} 
While methods mentioned in the previous section utilize the source data during adaptation, there has been a surge of interest in adaptation using only a pre-trained source model. Recent approaches in source-free adaptation have primarily focused on the image classification task. These include techniques such as information maximization~\cite{liang2020we,ahmed2021unsupervised}, pseudo-labeling~\cite{yeh2021sofa} and self-supervision~\cite{xia2021adaptive}. In the context of semantic segmentation, \cite{Liu_CVPR_2021} proposed an algorithm that combines ideas from the above methods by implementing a curriculum learning scheme. Similar ideas are explored in~\cite{wang2021domain}, where the authors additionally explore the concept of negative learning~\cite{kim2019nlnl}. Finally, in~\cite{S_2021_CVPR}, pseudo-labeling and uncertainty estimation via dropout are used to tackle the problem. In contrast to all these approaches which approach source-free adaptation in a \emph{uni-modal} setting, our framework performs source-free adaptation in a \emph{multi-modal} setting by exploiting the intrinsic semantic relationships between the different modalities. 
\vskip 6pt
\noindent \textbf{Cross-modal Domain Adaptation.} 
Despite the keen interest in multi-modal analysis, there have been few works that attempt this in the domain adaptive scenario. Recently, a cross-modal UDA (xMUDA) setting was proposed~\cite{jaritz2020xmuda}, where modalities learn from each other through mutual cycle-consistency to prevent the stronger modality from adopting false predictions from the weaker one. A more recent work \cite{ahmed2022cross}, motivated by the fact that the availability of multi-modal data and models is usually limited, considered adapting source model(s) trained on some source modality to a different target modality without access to the source data. Our problem setting also considers that uni-modal models are more widely available than multi-modal ones, but addresses a different problem: adapting uni-modal models of different modalities to a multi-modal target without source data. Critically different from~\cite{jaritz2020xmuda}, we do not assume the existence of paired multi-modal training data and do not need access to the source data during adaptation.

\section{Method} \label{sec:method}

\subsection{Problem Setting} \label{method:problem_setting}

We address the problem of adapting a set of models trained independently on uni-modal data to a target domain consisting of unlabeled, multi-modal data, without access to the original source dataset. We consider a set of source models, each of a unique modality, trained in a supervised manner on a set of $K$ categories. Once these models have been trained, the source data is discarded. We are then given a new target domain containing unlabeled, paired, multi-modal data, corresponding to each of the modalities present in the source datasets. We aim to adapt the source models to this new domain by exploiting the semantic relationships that inherently exist between the modalities. 

Throughout this work we will consider two input modalities: 2D image data $\mathcal{X}^{2D}$, of dimension $H\times W\times 3$, and 3D point cloud data $\mathcal{X}^{3D}$, of dimension $N\times 3$ ($N$ is the number of points), each of which has a corresponding label space $\mathcal{Y}^{2D}$ and $\mathcal{Y}^{3D}$. The source domain $\mathcal{D}_\mathcal{S}=\{\{x_i^{2D}, y_i^{2D}\}_{i=1}^{N_{S,2D}}, \{x_i^{3D}, y_i^{3D}\}_{i=1}^{N_{S,3D}}\}$ consists of samples $x^{2D}\in\mathcal{X}^{2D}$ and $x^{3D}\in\mathcal{X}^{3D}$ and their corresponding labels $y^{2D}\in\mathcal{Y}^{2D}$ and $y^{3D}\in\mathcal{Y}^{3D}$, with no known association between $x^{2D}$ and $x^{3D}$. We assume that each source model, $\mathcal{M}^{2D}:\mathcal{X}^{2D} \rightarrow \mathcal{Y}^{2D}$ and $\mathcal{M}^{3D}:\mathcal{X}^{3D} \rightarrow \mathcal{Y}^{3D}$, is trained with access to only a single input modality. The models can be decomposed as $\mathcal{M}^{2D} = f^{2D}\circ g^{2D}$ and $\mathcal{M}^{3D} = f^{3D}\circ g^{3D}$, where $f^{2D},f^{3D}$ are the respective feature encoders and $g^{2D},g^{3D}$ are the corresponding linear classifiers. We adapt these source models to the target domain $\mathcal{D}_\mathcal{T}=\{x_i^{2D}, x_i^{3D}\}_{i=1}^{N_T}$ consisting of paired samples  $x_i^{2D}\in \mathcal{X}^{2D}$ and $x_i^{3D}\in X^{3D}$, but no label information. 

\begin{figure*}[t]
    \centering
    \includegraphics[width=0.75\textwidth]{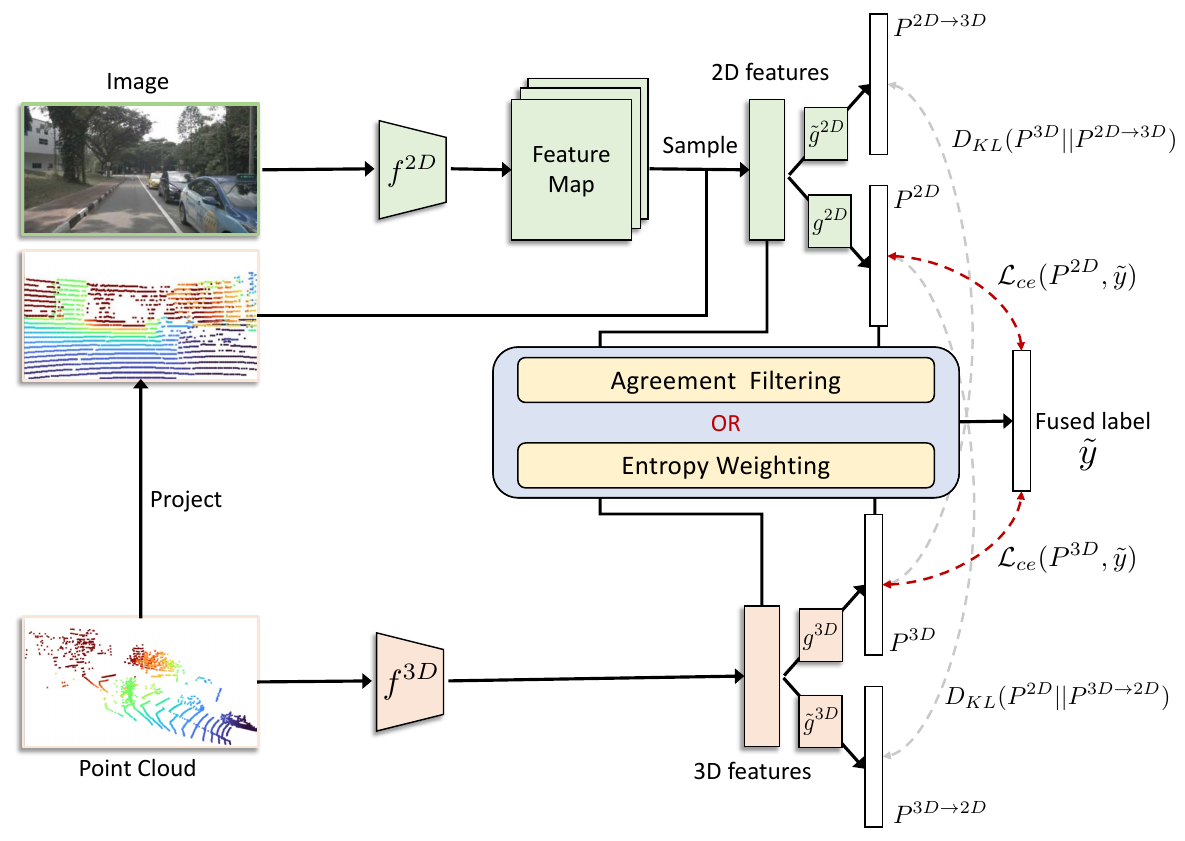}
    \caption{\textbf{Overview of framework.} Our framework consists of two streams corresponding to the 2D and 3D inputs respectively. The two modalities are processed by separate feature encoders $f^{2D} \text{and } f^{3D}$, with the 2D features being sampled by projecting the 3D points onto the corresponding RGB image. The four segmentation outputs consist of the main predictions $P^{2D},P^{3D}$ and the modality translated predictions $P^{2D\rightarrow 3D},P^{3D\rightarrow 2D}$ (using auxiliary heads $\tilde{g}^{2D},\tilde{g}^{3D}$). The main predictions are used to compute pseudo-labels via a fusion strategy, agreement filtering or entropy weighting. The choice of approach is decided via a switching strategy. These pseudo-labels are used to train the framework using cross-entropy loss. On the other hand, we transfer knowledge across modalities using $D_{KL}(P^{3D}||P^{2D\rightarrow 3D})$, where the objective of the 2D translation head is to estimate the main 3D output and vice versa, $D_{KL}(P^{2D}||P^{3D\rightarrow 2D})$.}
    \label{fig:overview}
\end{figure*}

\subsection{Framework Overview} \label{method:overview}
We adapt the source models via a pseudo-labeling approach, where predictions from the source models are fused across modalities to supervise the training on the unlabeled target dataset. Fusing pseudo-labels allows us to create more accurate labels for adaptation, and implicitly allows the usage of cross-modal correlations as part of the process. 

We propose a switching framework which automatically determines how to change between two complementary fusion methods: (i) \textbf{agreement filtering}, and (ii) \textbf{entropy weighting}. Agreement filtering utilizes the consensus between predictions from the different modalities as a score to combine and refine the pseudo-labels. On the other hand, entropy weighting employs an information-theoretic approach to fuse predictions across modalities, supplemented by a hypothesis testing over class-conditional feature space statistics to refine predictions. The optimal fusion approach is chosen \emph{automatically} during adaptation via a switching strategy which utilizes meta-data derived from the source models.

Even with refined pseudo-labels, adaptation is still challenging due to the fact that semantic categories in the real-world are usually quite imbalanced. Thus, we present a method for estimating the class distribution, which in turn can facilitate the learning process. Using the estimated class distributions, the refined pseudo-labels are subsequently used to jointly adapt the source models to the target domain. An overview of our framework is shown in Figure~\ref{fig:overview}.


\subsection{Pseudo-label Fusion} \label{method:fusion}
The pseudo-labels obtained from the uni-modal source models are potentially noisy due to the domain shift between the source and the target domains. Using these directly to conduct the adaptation process can degrade the performance of the adapted models. 
Thus, we propose a switching framework that automatically chooses between two complementary methods to fuse pseudo-labels across modalities in order to obtain a single pseudo-label to supervise adaptation.
\vskip 4pt
\noindent \textbf{Agreement Filtering.} 
In agreement filtering (AF), we implement fusion by comparing the pseudo-labels across modalities and keeping only those that agree with each other. First, we calculate pseudo-labels for each modality, by taking the most likely prediction according to the unadapted source model. Next, we perform an initial refinement by filtering out any pseudo-label with a confidence score that is lower than the median confidence value for that class, \emph{i.e.}
\begin{align}
\tilde{y}_i^{2D} &= \begin{cases}
                       \text{arg}\max_{k} \mathcal{M}^{2D}_k(x_i^{2D}), & \mathcal{M}^{2D}_k(x_i^{2D}) \geq \text{m}_k\\
                       \textrm{ignore},& otherwise
                   \end{cases} \label{eq:certainty_filter}\\
\text{m}_k &= \text{median}(\{\mathcal{M}^{2D}_k(x^{2D})|x^{2D}\in\mathcal{D}_\mathcal{T}\})
\end{align}
where $k$ denotes the class index. We use the median here since it tends to be a better indicator of the central tendency given skewed data. Once both $2D$ and $3D$ pseudo-labels are filtered, we combine them via an agreement filter as follows, 
\begin{align}
    \tilde{y}_i &= \begin{cases}
                    \tilde{y}_i^{2D},& \tilde{y}_i^{2D} = \tilde{y}_i^{3D}\\
                    \textrm{ignore},& \tilde{y}_i^{2D} \neq \tilde{y}_i^{3D}\\
                \end{cases}
\end{align}
Consequently, the new set of refined pseudo-labels contains only highly confident labels that agree across modalities. Certain types of bias will be specific to each modality. For example, a model trained on image data might associate green pixels with vegetation, but a model trained on point clouds does not sense color and cannot share this bias. Since we require every source model agree and each source model is trained on a different modality, we are left with a single common modality agnostic pseudo-label. 

\noindent \textbf{Entropy Weighting.} 
Entropy weighting (EW) explores an alternative information-theoretic method for combining pseudo-labels. Inspired by~\cite{shen2022benefits}, we perform an initial fusion by linearly combining the predictions from the source models by using uncertainty weights derived from the entropy of the output probabilities. Specifically,
\begin{align}
    p &= w^{2D}\psi(\mathcal{M}^{2D}(x^{2D})) + w^{3D}\psi(\mathcal{M}^{3D}(x^{3D}))\\
    w^{2D} &= \frac{e^{-h(\mathcal{M}^{2D}(x^{2D}))}}{e^{-h(\mathcal{M}^{2D}(x^{2D}))} + e^{-h(\mathcal{M}^{3D}(x^{3D}))}}\\
    w^{3D} &= 1 - w^{2D}
\end{align}
where $h(x)=-\sum_{k=1}^K\psi_k(x) log(\psi_k(x))$ is the entropy over the output space, $\psi$ is the softmax function, and $p$ denotes the merged output probabilities. The merged probabilities are filtered using the class median of the merged probability, as defined previously for individual modalities in Eq.~\eqref{eq:certainty_filter}, giving a single fused pseudo-label.
\vskip 4pt
\noindent \textbf{Entropy Weighting Refinement.} 
While this weighted combination is useful for resolving minor disagreements between modalities, it may also transfer noise across modalities leading to accurate pseudo-labels being rejected. In order to recover these pseudo-labels, we integrate target dataset statistics via hypothesis testing. The statistics we make use of are the class-wise mean and standard deviation calculated in the 2D and 3D feature spaces. We use the pseudo-labels previously accepted by Entropy Weighting as the class labels to calculate these statistics. The mean feature for class $k$ in the 2D feature space is notated as $\mu_k^{2D}$ and the corresponding standard deviation as $\sigma_k^{2D}$ and similarly we define $\mu_k^{3D}$ and $\sigma_k^{3D}$ for the 3D feature space.

Given a sample consisting of paired observations $x^{2D}$ and $x^{3D}$ and the unadapted source models $\mathcal{M}^{2D}$ and $\mathcal{M}^{3D}$ we calculate a hypothesis pseudo-label using each modality individually as
\begin{align}
    k_{2D} = \text{arg}\max_{k} \mathcal{M}^{2D}(x^{2D})\\
    k_{3D} = \text{arg}\max_{k} \mathcal{M}^{3D}(x^{3D}).
\end{align}
Within the 2D feature space, we take $k_{2D}$ as the null hypothesis and $k_{3D}$ as the alternative hypothesis. We assume the likelihood function of a hypothesis to be a multivariate normal distribution parameterized by the class-wise mean and variance. We can then perform a likelihood ratio test:
\begin{align}
\frac{\mathcal{N}(f^{2D}(x^{2D}); \mu_{k_{2D}}^{2D}, \big(\sigma_{k_{2D}}^{2D}\big)^2)}{\mathcal{N}(f^{2D}(x^{2D}); \mu_{k_{3D}}^{2D}, \big(\sigma_{k_{3D}}^{2D}\big)^2)} \leq \tau.
\end{align}
We set $\tau=1$ throughout our experiments and perform a sensitivity analysis of the threshold in the Appendix \ref{apdx:threshold}. We perform the same test in the 3D feature space, however in this case we take $k_{3D}$ as the null hypothesis and $k_{2D}$ as the alternative.

If both hypothesis tests agree on the correct hypothesis, e.g. the 2D feature space test rejects $k_{2D}$ in favor of $k_{3D}$ and the 3D feature space test keeps the $k_{3D}$, we take this as the correct pseudo-label. If the hypothesis tests do not agree on which hypothesis is correct, the pseudo-label is rejected. This allows us to recover previously rejected pseudo-labels and the entire set is used to supervise the adaptation process, as shown in Section \ref{method:training}.
\vskip 4pt
\noindent \textbf{Automatic Switching Between Fusion Methods.}
The performance of the two fusion techniques previously described varies depending on the particulars of the adaptation task. When the domain gap is small we expect EW to work better since it can resolve small disagreements across modalities, otherwise, AF is chosen since fusing discrete class labels has less potential to transfer noise. We cannot compute the domain gap directly, so we propose to measure it indirectly via the agreement between modalities. We justify the use of agreement as a proxy measure for domain gap via an empirical examination of agreement rates across various adaptation scenarios in Section~\ref{experiments:results}.

First, we estimate the expected rate of agreement between the unadapted source models. Models agree with each other either if they both predict the same correct class, or both predict the same incorrect class. Considering only the case where the models agree and are correct, we can lower bound the expected agreement rate using source meta-data, more specifically, the Top-1 accuracy of the model on the source data. \footnote{The Top-1 accuracy of a model is a statistic commonly reported for most models.} Considering the predictions from each model to be independent of each other, the expected agreement can be lower bounded by the product of the Top-1 accuracies, \emph{i.e.}
\begin{align}
    \Scale[0.9]{
    P(\hat{y}^{2D} = \hat{y}^{3D}) \geq \mathcal{P}(\hat{y}^{2D} = y|x^{2D})\mathcal{P}(\hat{y}^{3D}=y|x^{3D}) . }
\end{align}
This is the likelihood that source models agree given the source data distribution, which is our null hypothesis.

The actual agreement rate between the source models on the target dataset is calculated using AF, as described in Section~\ref{method:fusion}, and is the proportion of accepted pseudo-labels. This gives us the likelihood that the models agree given the target dataset distribution, which we take as the alternative hypothesis.

We compare these two likelihoods using hypothesis ratio testing. If the ratio between them is close to unity, then neither data distribution is more likely based on the agreement rate. Thus, we can infer that the target data comes from a data distribution similar to the source, and EW is chosen as the fusion method. The more the ratio diverges from unity the less likely the source and target datasets have a similar distribution. Throughout our experiments we set threshold of $0.5$ to automatically switch to AF. We examine the choice of threshold in Section \ref{experiments:results}

\subsection{Training}\label{method:training}
Using pseudo-labels, refined using AF or EW as chosen by our switching method, we optimize the following loss to adapt the uni-modal models to the target domain,
\begin{align}
    \mathcal{L}_{pl} = \mathcal{L}_{ce}(P^{2D}, \tilde{y}) + \mathcal{L}_{ce}(P^{3D}, \tilde{y}) .
\end{align}
Here, $\mathcal{L}_{ce}$ denotes the weighted cross-entropy loss, $P^{2D}=\psi(\mathcal{M}^{2D}(x^{2D}))$, and $P^{3D}=\psi(\mathcal{M}^{3D}(x^{3D}))$. We use the weighted cross-entropy in order to address potential class imbalances during adaptation, using the estimates of the class distribution as the weights. Note that in the source-free setting we do not have access to labels either on the source or on the target domain. Thus, we estimate the class distribution by taking the mean of the output distribution of each modality across target instances as $\tilde{p}^{2D}=\mathbb{E}_{\mathcal{D}_\mathcal{T}}\left[P^{2D}\right]$. We estimate $\tilde{p}^{3D}$ by averaging over the output of $\mathcal{M}^{3D}$ in a similar fashion. The final estimate of the class distribution is calculated by averaging $\tilde{p}^{2D}$ and $\tilde{p}^{3D}$ as
$\tilde{p} = \frac{\tilde{p}^{2D} + \tilde{p}^{3D}}{2}$. Inspired from~\cite{jaritz2020xmuda}, we optimize an additional cross-modal cycle-consistency constraint to allow information transfer between modalities through mutual imitation, \emph{i.e.}
\begin{align}
    \Scale[0.85]{
    \mathcal{L}_{xM} = D_{KL}(P^{2D}||P^{3D\rightarrow 2D}) + D_{KL}(P^{3D}||P^{2D\rightarrow 3D}).}
\end{align}
$P^{3D\rightarrow 2D}$ is an output head of $\mathcal{M}^{3D}$ which attempts to predict $P^{2D}$ and likewise $P^{2D\rightarrow 3D}$ is a prediction of $P^{3D}$ by $\mathcal{M}^{2D}$ (see Figure~\ref{fig:overview}). 
Minimizing the KL divergence $D_{KL}$ allows for information flow between the two models.
We combine these to get, $\mathcal{L}_{tot} = \mathcal{L}_{pl} + \lambda\mathcal{L}_{xM}$, where $\lambda$ is a hyper-parameter to balance the two loss functions and is typically set to $0.1$. More training details and pseudo-code can be found in the Appendix.
\section{Experiments}\label{experiments}
In this section, we demonstrate the ability of our proposed framework to perform source-free adaptation of uni-modal models to multi-modal data. Experiments across multiple adaptation scenarios demonstrate that the pseudo-label refinement across modalities, in addition to the cross-modal consistency criterion, provides a consistent improvement in both 2D and 3D target domain performance over baselines. Even in scenarios where the 2D and 3D source data are derived from distinct domains, our framework is able to adapt the disconnected models to multi-modal datasets effectively. 
\begin{table*}[ht]
    \centering
    \caption{\textbf{Results on primary adaptation scenarios.} Our framework outperforms both the uni-modal and multi-modal source-free adaptation baselines by a large margin across all scenarios. Note that xMUDA needs source data, and the results below are provided assuming that the source data is available during adaptation. Despite xMUDA requiring access to source data and models that are jointly trained on multi-modal data, our method has comparable performance and even outperforms it in the A2D2/SemanticKITTI scenario.} 
    \vskip 2pt
    \resizebox{\textwidth}{!}{
    \begin{tabular}{lccccccccccc}
      \toprule
       &  &  & \multicolumn{3}{c}{\textbf{USA/Singapore}} & \multicolumn{3}{c}{\textbf{Day/Night}} & \multicolumn{3}{c}{\textbf{A2D2/SemKITTI}}\\
      \cmidrule(lr){4-6}\cmidrule(lr){7-9}\cmidrule(lr){10-12}
      \textbf{Method} & \textbf{Multimodal} & \textbf{Source-free} & 2D & 3D & 2D+3D & 2D & 3D & 2D+3D & 2D & 3D & 2D+3D \\
      \midrule
      No adaptation & - & - & 49.67 & 47.39 & 55.96 & 41.85 & 42.32 & 49.98 & 32.85 & 34.01 & 40.52\\
      xMUDA & \cmark & \xmark & 61.10 & 54.10 & 63.20 & 47.10 & 46.70 & 50.80 & 43.70 & 48.50 &  49.10\\
      \midrule
      xMUDA SF & \cmark & \cmark & 51.88$\pm$0.12 & 51.04$\pm$0.27 & 56.59$\pm$1.46 & 36.43$\pm$1.05	& 42.23$\pm$1.62 & 44.46$\pm$1.49 & 33.74$\pm$0.15 & 36.79$\pm$0.23 & 42.68$\pm$0.19\\
      SHOT & \xmark & \cmark & 52.21$\pm$0.00 & 48.20$\pm$0.01 & 58.18$\pm$0.00 & 34.70$\pm$0.03	& 40.61$\pm$0.00 & 40.83$\pm$0.01 & 33.26$\pm$0.23	& 36.92$\pm$0.06 & 42.09$\pm$0.18\\
      \midrule
      SUMMIT (Ours) & \cmark & \cmark & \textbf{57.92$\pm$1.43}	 & \textbf{52.95$\pm$0.72} & \textbf{61.10$\pm$1.97} & \textbf{44.73$\pm$0.01}	& \textbf{44.53$\pm$0.00} & \textbf{50.73$\pm$0.01} & \textbf{44.68$\pm$0.30}	& \textbf{48.56$\pm$0.34} & \textbf{49.93$\pm$0.13}\\
      \midrule
      Oracle & - & - & 64.67 & 57.19 & 71.29 & 46.11 & 40.31 & 45.60 & 58.84 & 71.75 & 74.57\\
      \bottomrule
    \end{tabular}\label{tab:performance}
}
\end{table*}
\subsection{Experimental Setup} \label{experiments:setup}
\noindent\textbf{Datasets and Experimental Scenarios.}
We use three autonomous vehicle datasets in our experiments, which are organized into three different source-free cross-modal UDA adaptation scenarios following the experimental setup defined in \cite{jaritz2020xmuda}. The critical difference is that our experiments have no access to the source data, and the source models are trained separately. Two of the adaptation experiments are derived from the \emph{nuScenes} dataset~\cite{caesar2020nuscenes}. The nuScenes dataset gathers data in multiple cities, throughout different times of day with multiple cameras and LIDAR. We extract approximately 34K images and the corresponding points clouds, and define two adaptation scenarios. The \emph{first scenario} takes all image and point clouds that were gathered in the USA as the source domain and those gathered in Singapore as the target domain. The \emph{second scenario} treats samples gathered during the day as the source domain and those gathered during the night as the target domain. 

The \emph{third scenario} considers two unrelated datasets as the source and target domain. The source domain is the \emph{A2D2} dataset~\cite{geyer2020a2d2}, which consists of approximately 40K images and point clouds. The target domain is the \emph{SemanticKITTI} dataset~\cite{behley2019semantickitti}, containing approximately 18K samples. The classes provided are dissimilar, thus, they are merged into larger super classes which are kept consistent between the domains; the details of this process can be found in the Appendix \ref{apdx:pseudo_code}. This scenario is particularly challenging due to the different sensors used in each dataset, in addition to the temporal and geographic shifts. 

While these experiments make no explicit use of pairing information when training the source models, there is the possibility of implicitly learning some cross-modal correlations since 2D and 3D samples are drawn from the same split of the dataset. Thus, we propose additional experiments, termed \emph{crossover experiments}, with no corresponding data points shared between the 2D and 3D datasets. We take the source/target splits defined on the nuScenes dataset and use each to train a single modality, \emph{i.e.} we use images from the USA split to train the 2D model and the point clouds from the Singapore split to train the 3D model. The SemanticKITTI dataset is used as the target dataset for all experiments. To increase the overlap of class labels between nuScenes and SemanticKITTI, we include additional labels for background objects from the  \emph{nuScenes-lidarseg} dataset~\cite{fong2021panoptic}. 
\vskip 2pt
\noindent\textbf{Baselines.} 
We compare our method to (i) \textbf{xMUDA}~\cite{jaritz2020xmuda}, which is a cross-modal adaption method, (ii) a version of xMUDA trained with only the data available in our setting called \textbf{xMUDA-SF}, and (iii) \textbf{SHOT}~\cite{liang2020we} which is one of the seminal works in source-free UDA. Note that xMUDA requires source data, and the results are provided assuming that the source data is available during adaptation. In xMUDA-SF we remove any loss functions that work on the source data and begin the adaptation process with independently trained source models. The performance of xMUDA-SF highlights the reliance of xMUDA on paired source data. To apply SHOT, we treat each modality as its own uni-modal UDA problem. We use this comparison to show the potential of modeling and exploiting cross-correlations. Each of the baseline methods uses the unrefined uni-modal pseudo-labels. In addition, we compare with the unadapted source models as a lower bound on performance (\textbf{no adaptation}), and models adapted using oracle ground truth labels as an upper bound (\textbf{Oracle}). We keep the same baselines for the crossover experiments, but we do not compare them to the original xMUDA method since no source pairing information exists, which xMUDA neeeds. We do not compare against multi-source source-free adaptation methods such as \cite{ahmed2021unsupervised}, because they assume that each source comes from the same modality. 
\vskip 2pt
\noindent\textbf{Experimental Objectives.}
The first three adaptation scenarios described above demonstrate that our method does not need access to paired source data, compared to state-of-the-art methods like~\cite{jaritz2020xmuda}, and that it can work with uni-modal trained models as the source. Furthermore, by comparing to existing source-free UDA methods, we will highlight the need to explicitly model cross-modal correlations.

Through the crossover experiments, we further show the reliance of \cite{jaritz2020xmuda} on \emph{paired} data. Critically, we demonstrate our method works with truly independent source models, opening up the possibility of adapting pretrained uni-modal models together to work on multi-modal data.
\vskip 2pt
\noindent\textbf{Implementation Details.}
Architectures for the 2D and 3D models follow from~\cite{jaritz2020xmuda}. Specifically, the 2D network is a UNet architecture~\cite{ronneberger2015u} with a ResNet34~\cite{he2016deep} backbone. To associate points across modalities, the point cloud is projected onto the image plane, and the corresponding features are sampled. These are then passed through a final classification layer to give us predictions. The 3D model uses the official PyTorch implementation of the Sparse Convolutional Networks UNet~\cite{graham20183d}. 
The source models are initially trained on each modality independently of each other for 100K iterations. 
Once the source model is trained, we save this model and use it as the same starting point for all adaptation methods to ensure a fair comparison. 

$\mathcal{L}_{xM}$ requires each model to provide a secondary output that predicts the output of the other modality. We initialize this second output head with a copy of the weights from the original output head. We set the $\lambda$ for $\mathcal{L}_{xM}$ according to the values used in \cite{jaritz2020xmuda}, and the initial learning rate to $1e-6$. Pseudo-labels are calculated offline before adaptation and are kept fixed throughout the adaptation process. Each method adapts the model for an additional 100K iterations. More details can be found in the Appendix \ref{apdx:hyper_parameters}.


\begin{table*}[ht]
    \centering
    \caption{\textbf{Results on crossover adaptation scenarios.} Differences in sensor configuration and label distribution result in all methods experiencing a drop in performance. While our method is not an exception to this, we obtain consistent improvements of up to 12\% over the unadapted source model and other adaptation approaches.} 
    \vskip 2pt
    \resizebox{\textwidth}{!}{
    \begin{tabular}{lcccccccccccccc}
      \toprule
       &  &  & \multicolumn{3}{c}{\textbf{(USA-2D,Singapore-3D)/SemKITTI}} & \multicolumn{3}{c}{\textbf{(Singapore-2D,USA-3D)/SemKITTI}} & \multicolumn{3}{c}{\textbf{(Day-2D,Night-3D)/ SemKITTI}} & \multicolumn{3}{c}{\textbf{(Night-2D,Day-3D)/SemKITTI}}\\
      \cmidrule(lr){4-6}\cmidrule(lr){7-9}\cmidrule(lr){10-12}\cmidrule(lr){13-15}
      \textbf{Method} & \textbf{Multimodal} & \textbf{Source-free} & 2D & 3D & 2D+3D & 2D & 3D & 2D+3D & 2D & 3D & 2D+3D & 2D & 3D & 2D+3D \\
      \midrule
      No adaptation & - & - & 18.98 & 28.94 & 26.77 & 20.22 & 28.45 & 25.51 & 24.21 & 28.54 & 29.96 & 16.14 & 26.60 & 20.52\\
      \midrule
      xMUDA SF & \cmark & \cmark & 20.02$\pm$0.52 & 32.68$\pm$1.01 & 28.10$\pm$0.65 & 20.05$\pm$0.28 & 33.64$\pm$1.10 & 28.21$\pm$0.03 & 24.64$\pm$0.14 & 34.01$\pm$0.10 & 29.54$\pm$0.17 & 16.13$\pm$0.03 & 27.97$\pm$0.03 & 25.88$\pm$0.03\\
      SHOT & \xmark & \cmark & 19.48$\pm$0.04 & 34.07$\pm$0.06 & 27.39$\pm$0.03 & 20.40$\pm$0.04 & 32.39$\pm$0.02 & 28.14$\pm$0.05 & 24.55$\pm$0.07 & 34.08$\pm$0.03 & 29.13$\pm$0.07 & 16.11$\pm$0.02 & 28.10$\pm$0.02 & 25.83$\pm$0.03\\
      \midrule
      SUMMIT (Ours) & \cmark & \cmark & \textbf{28.84$\pm$0.72} & \textbf{33.42$\pm$0.73} & \textbf{31.53$\pm$0.63} & \textbf{28.74$\pm$0.70} & \textbf{34.07$\pm$1.01} & \textbf{31.41$\pm$0.47} & \textbf{32.19$\pm$0.24} & \textbf{38.71$\pm$0.27} & \textbf{35.51$\pm$0.16} & \textbf{28.51$\pm$0.16} & \textbf{32.83$\pm$0.14} & \textbf{30.78$\pm$0.13}\\
      \midrule
      Oracle & - & - & 55.36 & 68.73 & 69.82 & 56.37 & 67.69 & 70.60 & 58.20 & 67.82 & 70.82 & 56.41 & 68.69 & 69.90 \\
      \bottomrule 
    \end{tabular}\label{tab:crossover_performance}
}
\end{table*}

\subsection{Results}\label{experiments:results}
We report the results on the three primary source-free adaptation scenarios in Table~\ref{tab:performance}. We report the mIoU for each modality, 2D and 3D, along with the average across the softmax output of both modalities, as 2D+3D. Unsurprisingly, the highest performance is achieved by xMUDA, \emph{however, this utilizes the source data during adaptation}. Its source-free counterpart, xMUDA-SF, obtains a much more modest improvement over the unadapted source, and in many cases leads to worse performance compared to no adaptation. This clearly highlights the reliance of xMUDA on the pairing information being present when training the source model. The improvement of SHOT is more consistent than xMUDA-SF, however since it is a uni-modal method, its improvement is limited when compared to our method.

Our framework achieves high-performance improvements over the source models across the board, with an improvement as high as 12\% over the other source-free methods. We see this in the A2D2/SemanticKITTI experiment, with an improvement of 12\% on individual modalities compared to xMUDA-SF and SHOT which provides only a few points of improvement. On the 2D+3D we have a smaller, but still significant, improvement of 7\% over SHOT. We also note that in several cases we are within 1\% of the original xMUDA score, which assumes access to the source dataset.

In Table~\ref{tab:crossover_performance}, we report the performance on the \emph{crossover experiments}. Note that SHOT outperforms xMUDA-SF in several cases, indicating that in the crossover experiments, xMUDA-SF tends to transfer more noise between the models. This may be due to lower performing source models since this adaptation scenario goes between two different datasets as opposed to the previous experiments where two of the experiments transfer between partitions of a single dataset. Our framework outperforms existing baselines, with improvements ranging from 4\%-10\% over the unadapted source. The largest improvement is on Night 2D \& Day 3D, despite the poor performance of the initial 2D source model. In some cases the performance of SHOT on the 3D modality approaches our performance; however, our method achieves a stronger improvement on the 2D and 2D+3D modality, showing that our method transfers useful information from the 3D to the 2D modality without transferring noise.

\begin{table}[t]
    \centering
    \caption{\textbf{Estimated source and target agreement.} The ratio between source and target agreement is used to estimate the domain gap. In the first two cases, which are partitions of the same data, the ratio is nearly one, so EW is selected. A lower ratio, as in the last case, indicates a high domain gap, so AF is selected.} 
    \vskip 2pt
    \resizebox{\columnwidth}{!}{
    \begin{tabular}{lccc}
      \toprule
       &  \textbf{USA/Singapore} & \textbf{Day/Night} & \textbf{A2D2/SemKITTI}\\
      \midrule
      Source Agreement & 94.25 & 95.33 & 84.75\\
      Target Agreement & 88.50 & 87.40 & 26.00\\
      \midrule 
      Ratio & 0.94 & 0.92 & 0.31\\
      \bottomrule
    \end{tabular}\label{tab:switching}
}
\end{table}
\begin{table}[t]
    \centering
    \small
    \caption{\textbf{Pseudo-label accuracy.} EW consistently admits a higher proportion of correct pseudo-labels than AF. However, in the case of a large domain gap, such as A2D2/SemanticKITTI, EW accepts a large amount of incorrect labels, which AF filters out.}
    \vskip 2pt
    \resizebox{\columnwidth}{!}{
    \begin{tabular}{*{10}{lccccccccc}}
      \toprule
       & \multicolumn{3}{c}{\textbf{USA/Singapore}} & \multicolumn{3}{c}{\textbf{Day/Night}} & \multicolumn{3}{c}{\textbf{A2D2/SemKITTI}}\\
      \cmidrule(lr){2-4}\cmidrule(lr){5-7}\cmidrule(lr){8-10}
      \textbf{Method} & Correct & Incorrect & Ignore & Correct & Incorrect & Ignore & Correct & Incorrect & Ignore \\
      \midrule
      AF & 87.29 & 0.13 & 12.58 & 88.45 & 0.10 & 11.46 & 23.76 & 1.73 & 74.51\\
      EW & 93.40 & 0.65 & 5.94 & 94.51 & 0.56 & 4.93 & 55.42 & 24.72 & 19.85\\
      \bottomrule
    \end{tabular}\label{tab:filtering}
}
\end{table}
\vskip 2pt
\noindent\textbf{Analysis of Automatic Switching.}
To gain further insight into the performance of our switching method, we analyze when particular pseudo-label fusion approaches are chosen, and the resulting quality of pseudo-labels that are being selected. In Table~\ref{tab:switching} we show the source and target agreement calculated according to the approach described in Section~\ref{method:fusion}, and the ratio between them on the first three experiments. For the USA/Singapore and Day/Night experiments, which are both derived by partitioning a single dataset, we obtain a ratio close to one and EW is selected. For the A2D2/SemanticKITTI experiment, which adapts across datasets, the ratio drops to $0.31$ which is less than our threshold of $0.5$ leading to AF being selected. 

The efficacy of the switching is shown by the pseudo-label accuracy in Table~\ref{tab:filtering}. In the first two cases, EW gives correct pseudo-labels for over 90\% of samples, while allowing less than 1\% of incorrect labels. In the third experiment AF is selected and while far fewer correctly labeled samples are accepted the amount of incorrect labels is also much lower ($<2\%$), i.e., AF selects fewer, but correct, samples. Throughout the \emph{crossover experiments}, AF is selected, a full analysis of which is included in the Appendix \ref{apdx:switching}.
\begin{table}[t]
    \centering
    \caption{We examine the interplay between AF, EW, $\mathcal{L}_{xM}$. We note that both AF and EW on their own provide an improvement across modalities and work well with the cross-modal loss. The cross-modal loss provides the most improvement in the 2D+3D performance.}
    \vskip 2pt
    \resizebox{\columnwidth}{!}{
    \begin{tabular}{lccccccccc}
      \toprule
       & \multicolumn{3}{c}{USA/Singapore} & \multicolumn{3}{c}{Day/Night} & \multicolumn{3}{c}{A2D2/SemKITTI}\\
      \cmidrule(lr){2-4}\cmidrule(lr){5-7}\cmidrule(lr){8-10}
      Method & 2D & 3D & 2D+3D & 2D & 3D & 2D+3D & 2D & 3D & 2D+3D \\
      \midrule
      No Adaptation & 49.67 & 47.39 & 55.96 & 41.85 & 42.32 & 49.98 & 32.85 & 34.01 & 40.52 \\
      \midrule
      AF & 55.73 & 53.25 & 58.48 & 38.68 & 41.38 & 39.41 & 44.92 & 48.54 & 50.05\\
      AF+$\mathcal{L}_{xM}$ & 56.78 & 53.29 & 58.83 & 37.78 & 41.36 & 39.10 & 45.01 & 48.34 & 49.60\\
      \midrule
      EW & 57.32 & 51.97 & 62.28 & 44.23 & 44.61 & 50.72 & 35.93 & 43.16 & 41.23 \\
      EW+$\mathcal{L}_{xM}$ & 57.47 & 52.12 & 62.32 & 44.73 & 44.53 & 50.72 & 36.82 & 43.26 & 41.24\\
      \midrule
      Oracle & 64.67 & 57.19 & 71.29 & 46.11 & 40.31 & 45.60 & 58.84 & 71.75 & 74.57\\
      \bottomrule
    \end{tabular}}\label{tab:ablation}
\end{table}

\begin{table}[t]
    \centering
    \caption{We examine the interplay between agreement filtering (AF), entropy weighting (EW), and the cross-modal loss $\mathcal{L}_{xM}$ for the \textit{crossover} experiments. We note that in the \textit{crossover} experiments AF provides substantial improvements, while EW does not. Also worth noting is that most of the performance gain comes from just using AF, with $\mathcal{L}_{xM}$ providing modest improvements.} 
    \vskip 2pt
    \resizebox{\columnwidth}{!}{
    \begin{tabular}{lcccccccccccc}
      \toprule
       &  \multicolumn{3}{c}{USA-2D,Singapore-3D} & \multicolumn{3}{c}{Singapore-2D,USA-3D} & \multicolumn{3}{c}{Day-2D,Night-3D} & \multicolumn{3}{c}{Night-2D,Day-3D}\\
      \cmidrule(lr){2-4}\cmidrule(lr){5-7}\cmidrule(lr){8-10}\cmidrule(lr){11-13}
      Method & 2D & 3D & 2D+3D & 2D & 3D & 2D+3D & 2D & 3D & 2D+3D & 2D & 3D & 2D+3D\\
      \midrule
      No adaptation & 18.98 & 28.94 & 26.77 & 20.22 & 28.45 & 25.51 & 24.21 & 28.54 & 29.96 & 16.14 & 26.60 & 20.52\\
      \midrule
      AF & 29.34 & 33.00 & 31.86 & 28.50 & 34.59 & 31.32 & 31.76 & 38.19 & 35.07 & 28.22 & 32.61 & 30.55\\
      AF+$\mathcal{L}_{xM}$ & 28.02 & 34.27 & 30.81 & 29.55 & 32.91 & 31.95 & 32.42 & 38.54 & 35.67 & 28.63 & 32.98 & 30.87\\
      \midrule
      EW & 16.39 & 20.38 & 18.41 & 16.20 & 17.87 & 17.22 & 19.61 & 18.93 & 20.60 & 7.27 & 7.84 & 8.23\\
      EW+$\mathcal{L}_{xM}$ & 15.64 & 18.57 & 16.88 & 16.26 & 19.49 & 18.08 & 19.64 & 19.33 & 20.68 & 7.71 & 7.97 & 8.60\\
      \midrule
      Oracle & 55.36 & 68.73 & 69.82 & 56.37 & 67.69 & 70.60 & 58.20 & 67.82 & 70.82 & 56.41 & 68.69 & 69.90 \\
      \bottomrule
    \end{tabular}}\label{tab:crossover_ablation}
\end{table}
\vskip 2pt
\noindent\textbf{Ablation Study.}
We perform an ablation study on the adaptation and the \emph{crossover experiments} where we examine the performance of each of our complementary pseudo-label fusion methods both with and without the cross-modal loss. In 
Table \ref{tab:ablation} we can clearly see the majority of improvements coming from the pseudo label fusion, with some additional improvements from the cross-modal loss. We note in the USA/Singapore and Day/Night experiments entropy weighting outperforms the agreement filtering method, due to the smaller domain gap. However, in this case, agreement filtering still gives strong performance. In the A2D2/SemanticKITTI agreement filtering performs better, since there is a larger domain gap. We also note that our switching method consistently picks the best method for pseudo-label fusion. 

\begin{figure}[t]
    \centering
    \captionsetup[subfigure]{labelformat=empty}
    \subfloat[(b) No Adaptation]{
        \subfloat[2D]{
            \includegraphics[width=0.3\linewidth]{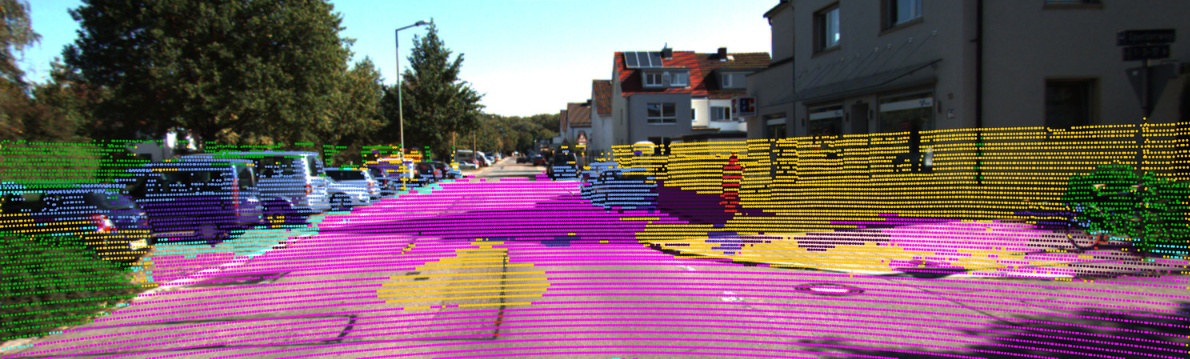}
        }
        \subfloat[3D]{
            \includegraphics[width=0.3\linewidth]{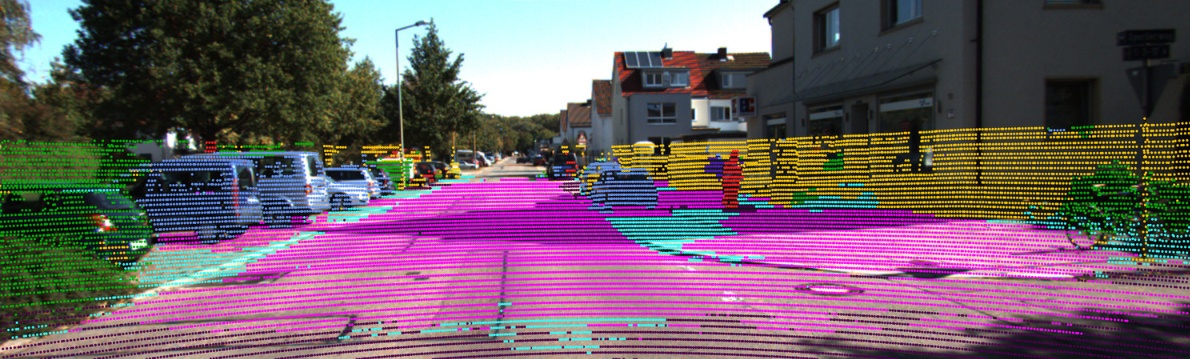}
        }
        \subfloat[2D+3D]{
            \includegraphics[width=0.3\linewidth]{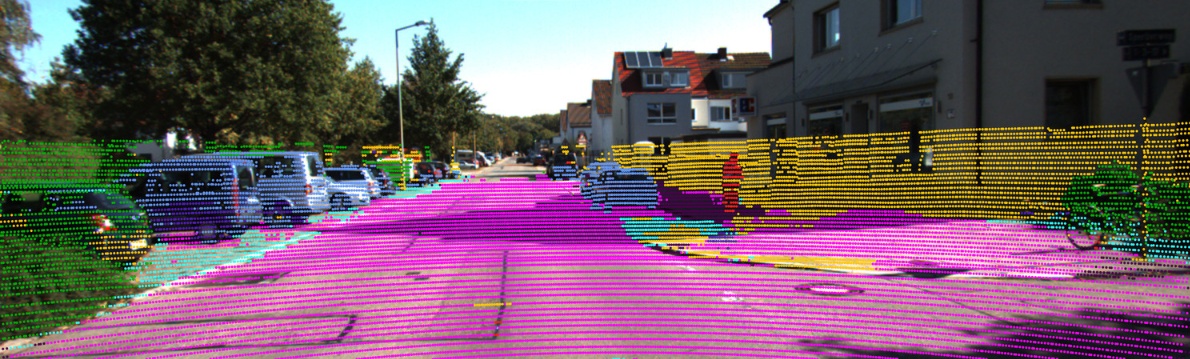}
        }
    }\\
    \vspace{-1em}
    \subfloat[(d) xMUDA Source Free]{
        \subfloat[2D]{
            \includegraphics[width=0.3\linewidth]{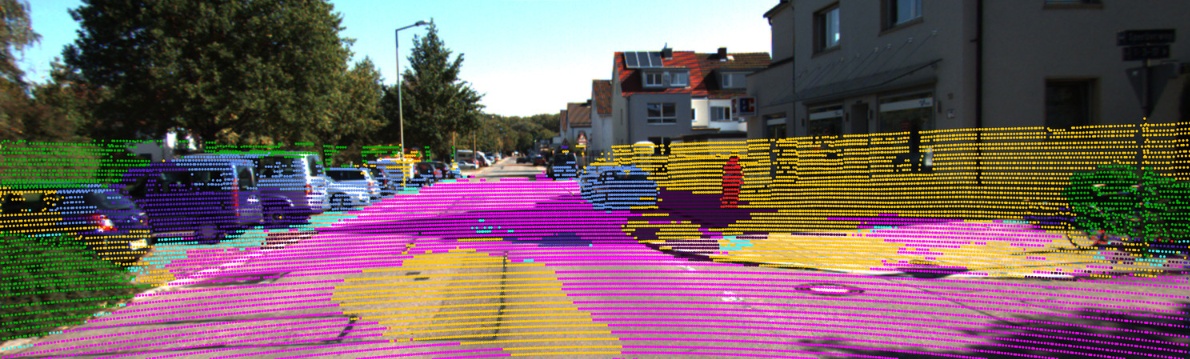}
        }
        \subfloat[3D]{
            \includegraphics[width=0.3\linewidth]{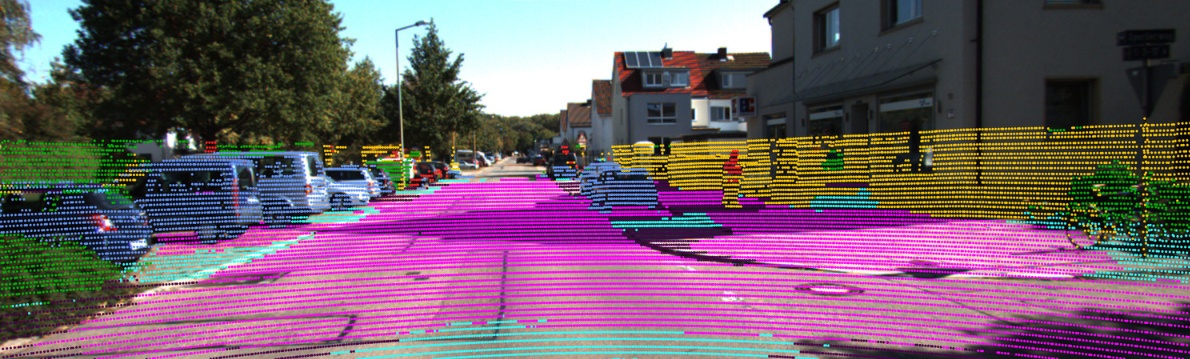}
        }
        \subfloat[2D+3D]{
            \includegraphics[width=0.3\linewidth]{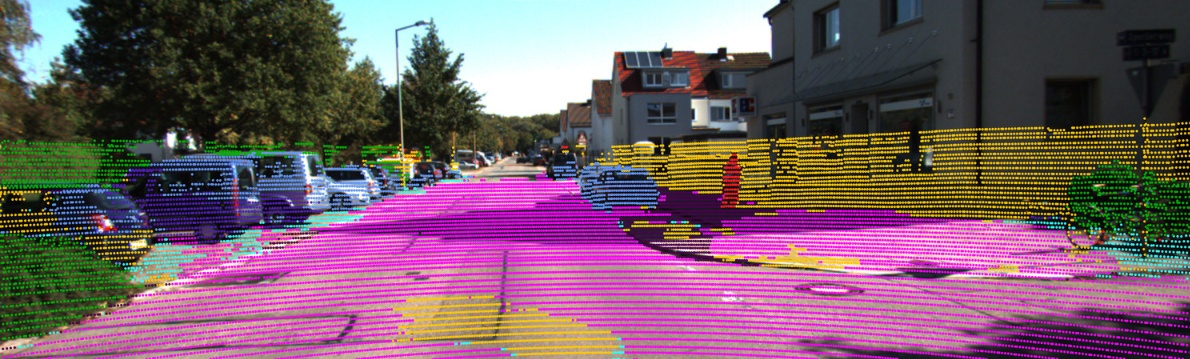}
        }
    }\\
    \vspace{-1em}
    \subfloat[(c) SUMMIT (Ours)]{
        \subfloat[2D]{
            \includegraphics[width=0.3\linewidth]{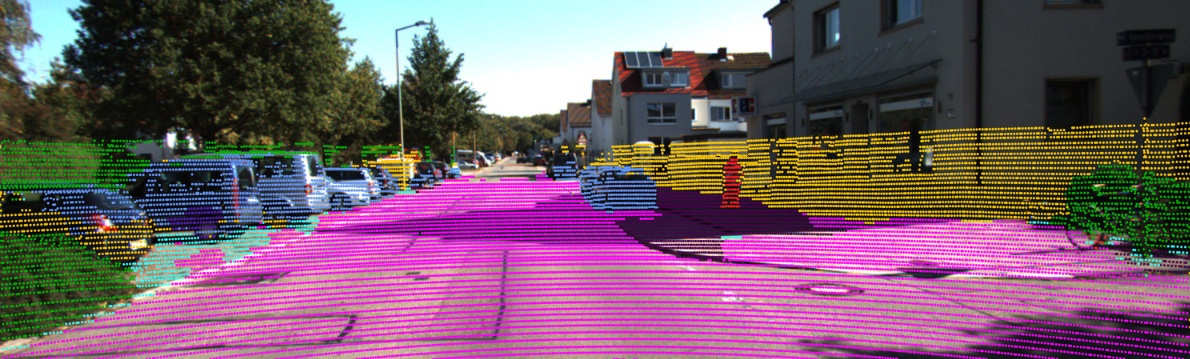}
        }
        \subfloat[3D]{
            \includegraphics[width=0.3\linewidth]{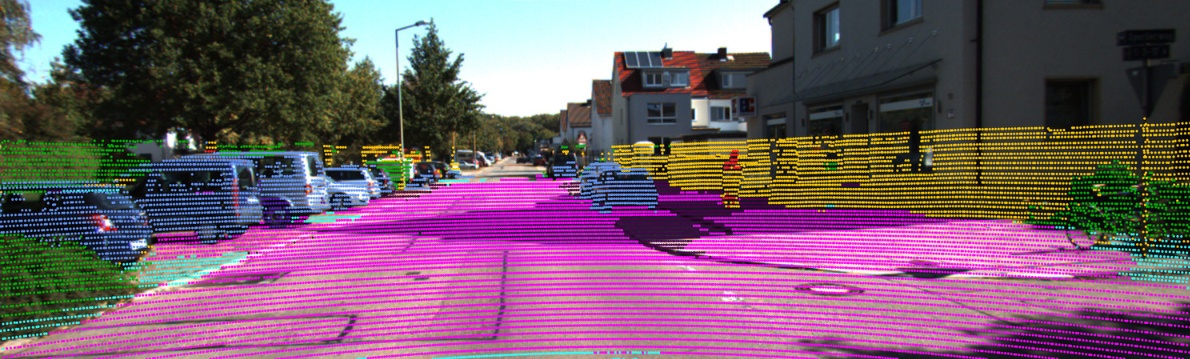}
        }
        \subfloat[2D+3D]{
            \includegraphics[width=0.3\linewidth]{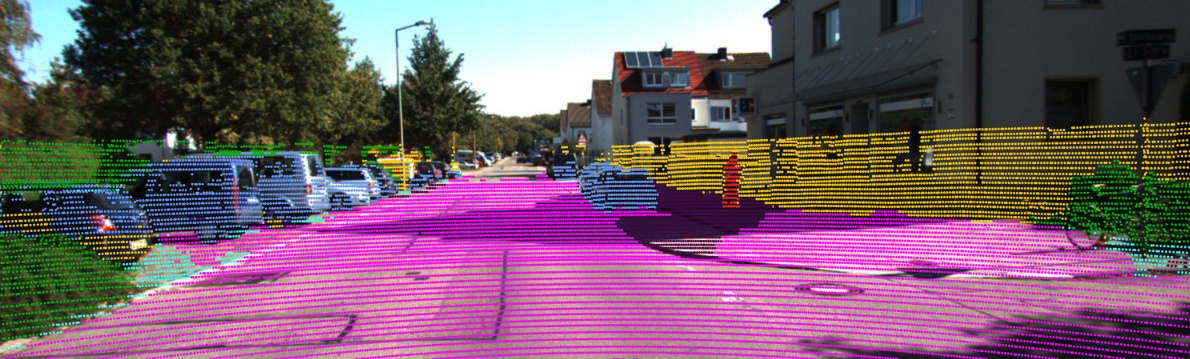}
        }
    }
    \caption{Qualitative results on the Day/Night to SemanticKITTI \textit{crossover} experiments. Once again we see xMUDA SF exacerbating misclassifications. Images are best viewed digitally.}
    \label{fig:nuscenes_lidarseg_semantic_kitti_plot}
\end{figure}

In Table \ref{tab:crossover_ablation} we show the results on the \emph{crossover experiments}. We see that AF far outperforms EW in this setting. Since the \emph{crossover} experiments are all across datasets, the domain gap is larger and AF is consistently selected. Most of the performance increase still comes solely from our pseudo-label fusion. We still see minor improvements from the cross-modal loss, but in some cases, this comes at the expense of one of the modalities.
\begin{figure}[t]
    \centering
    \includegraphics[width=\linewidth]{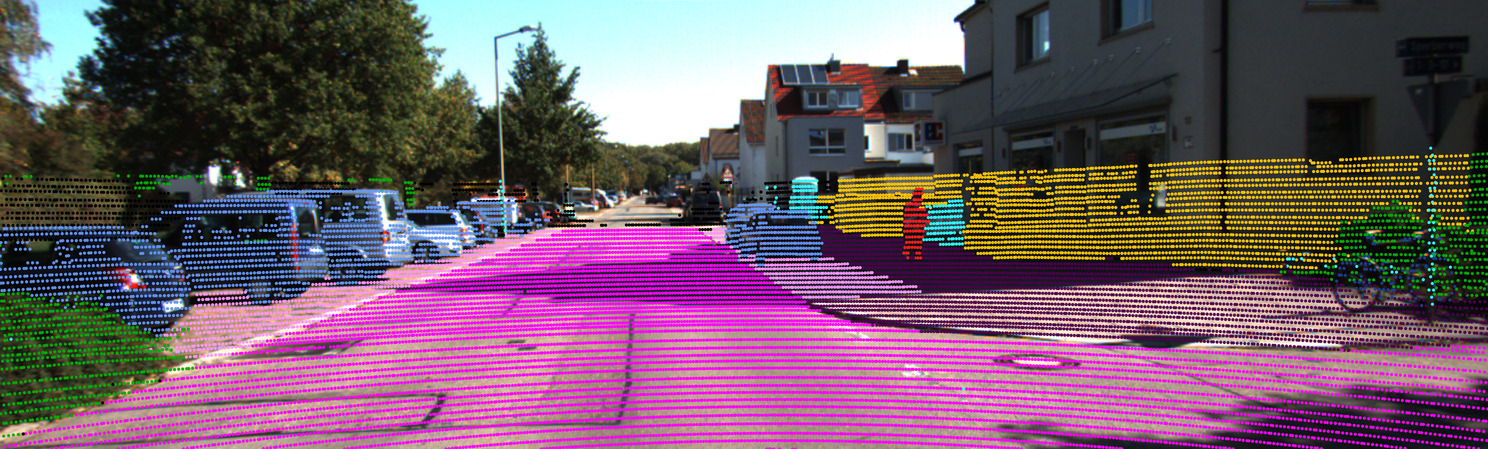}
    \caption{Ground truth data for comparison with qualitative results. Each color represents a different segmentation class.}
    \vspace{-2em}
    \label{fig:semantic_kitti_ground_truth}
\end{figure}
\begin{figure}[t]
    \centering
    \captionsetup[subfigure]{labelformat=empty}
    \subfloat[(b) No Adaptation]{
        \subfloat[2D]{
            \includegraphics[width=0.3\linewidth]{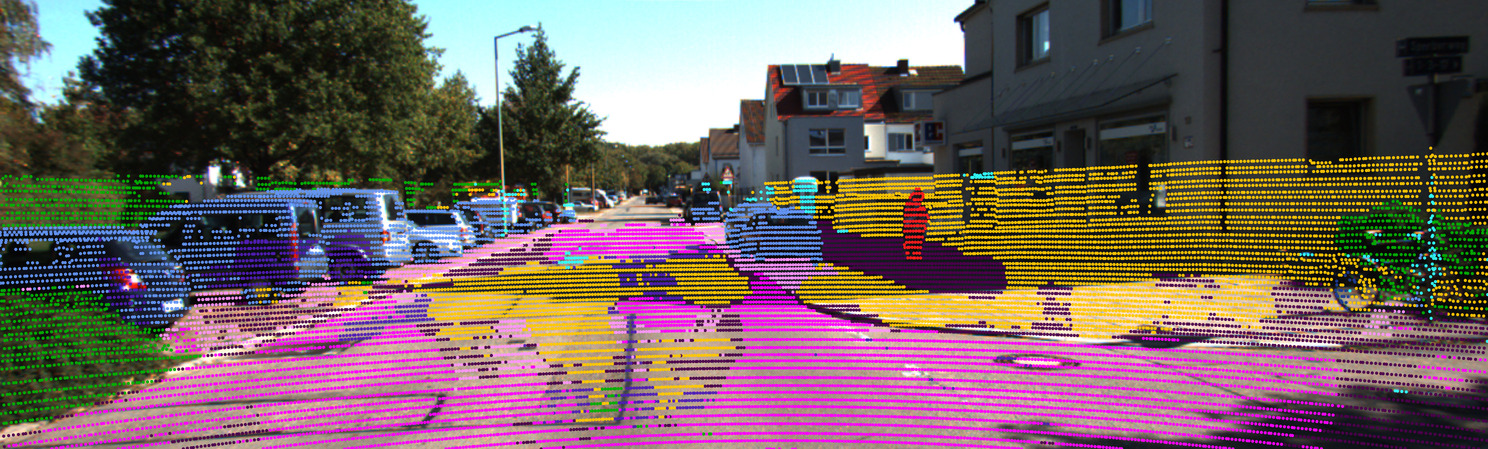}
        }
        \subfloat[3D]{
            \includegraphics[width=0.3\linewidth]{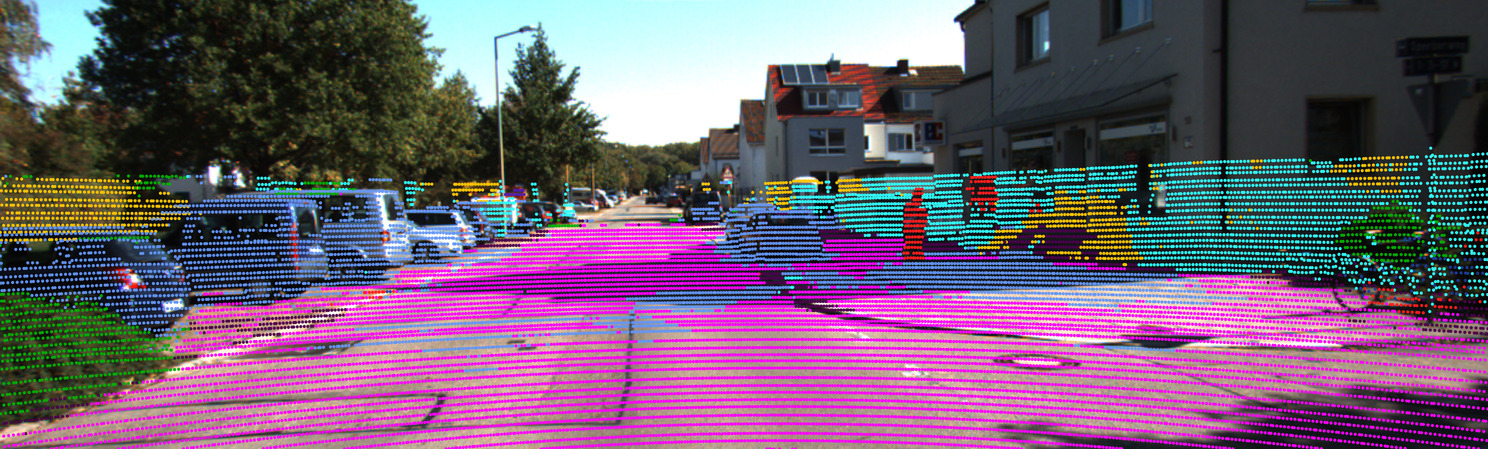}
        }
        \subfloat[2D+3D]{
            \includegraphics[width=0.3\linewidth]{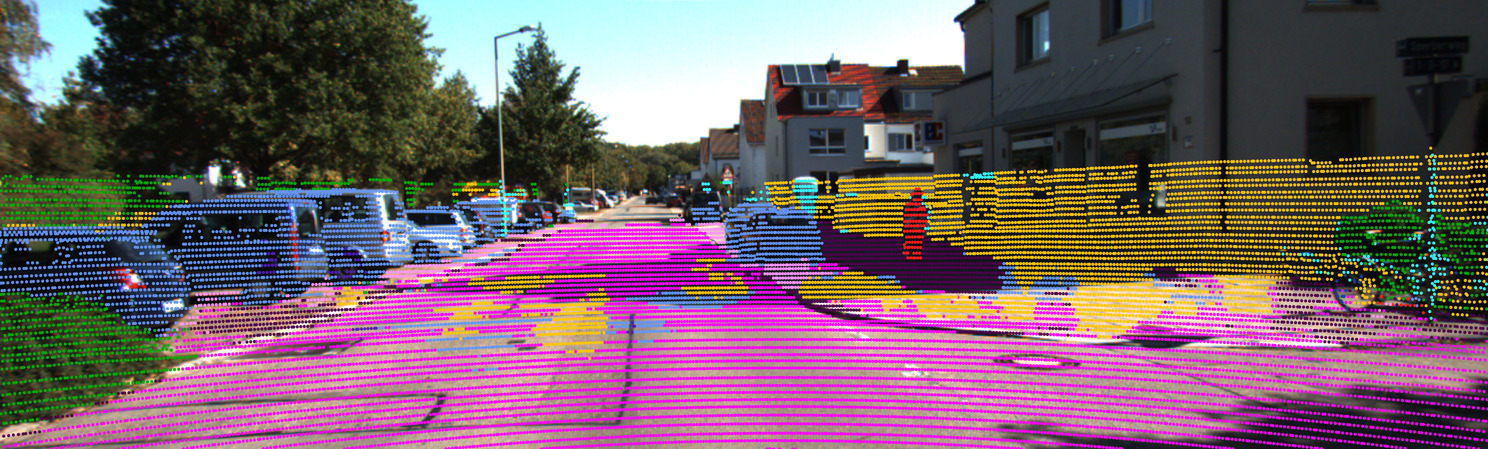}
        }
    }\\
    \vspace{-1em}
    \subfloat[(d) xMUDA Source Free]{
        \subfloat[2D]{
            \includegraphics[width=0.3\linewidth]{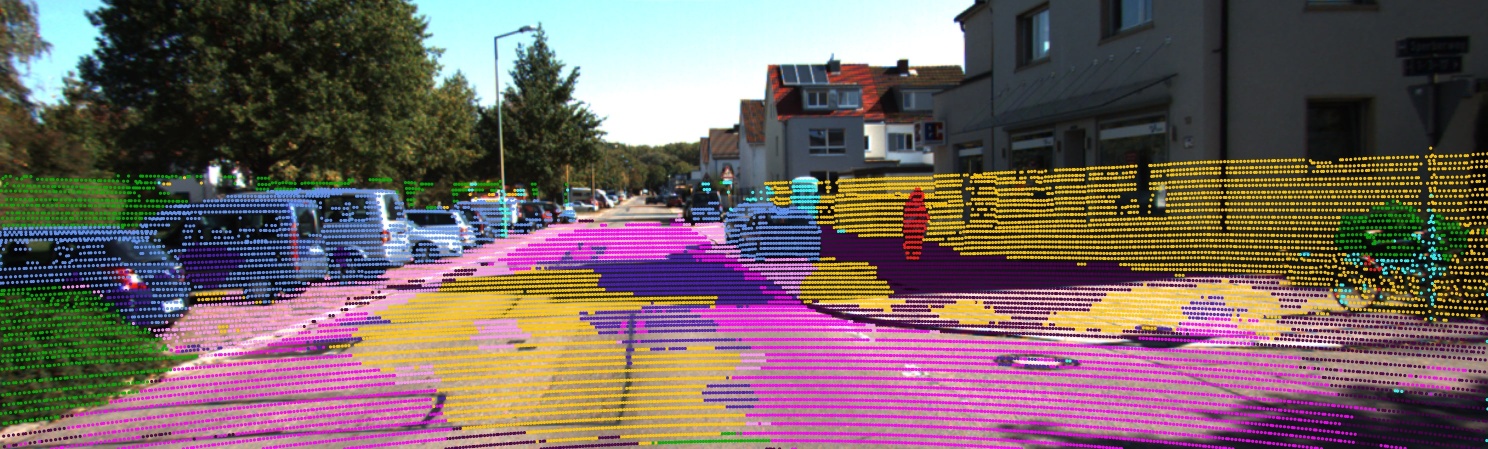}
        }
        \subfloat[3D]{
            \includegraphics[width=0.3\linewidth]{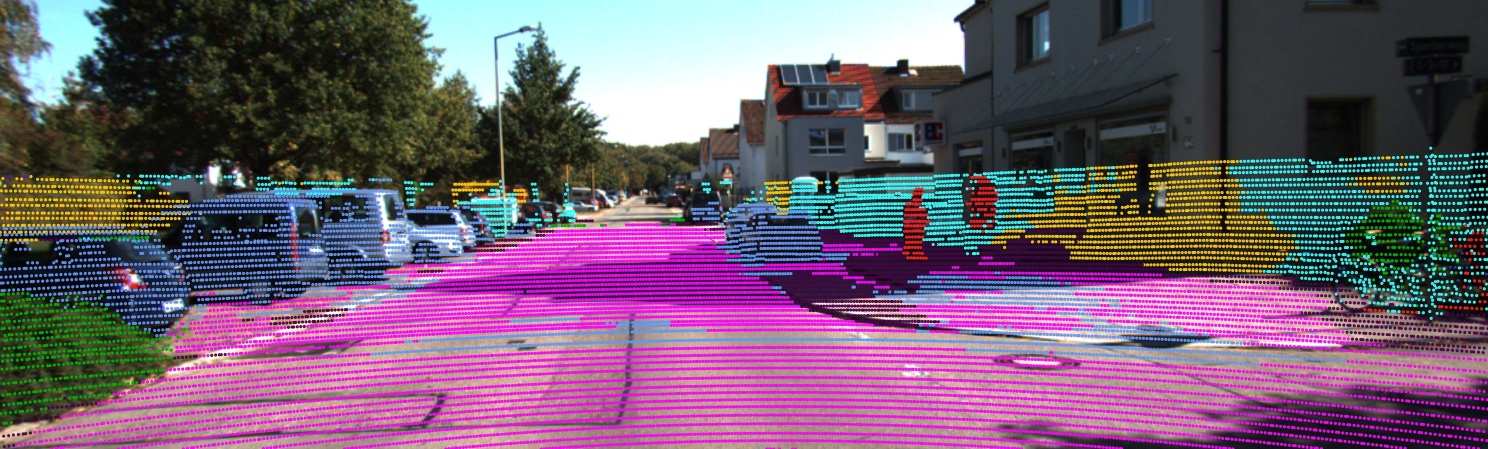}
        }
        \subfloat[2D+3D]{
            \includegraphics[width=0.3\linewidth]{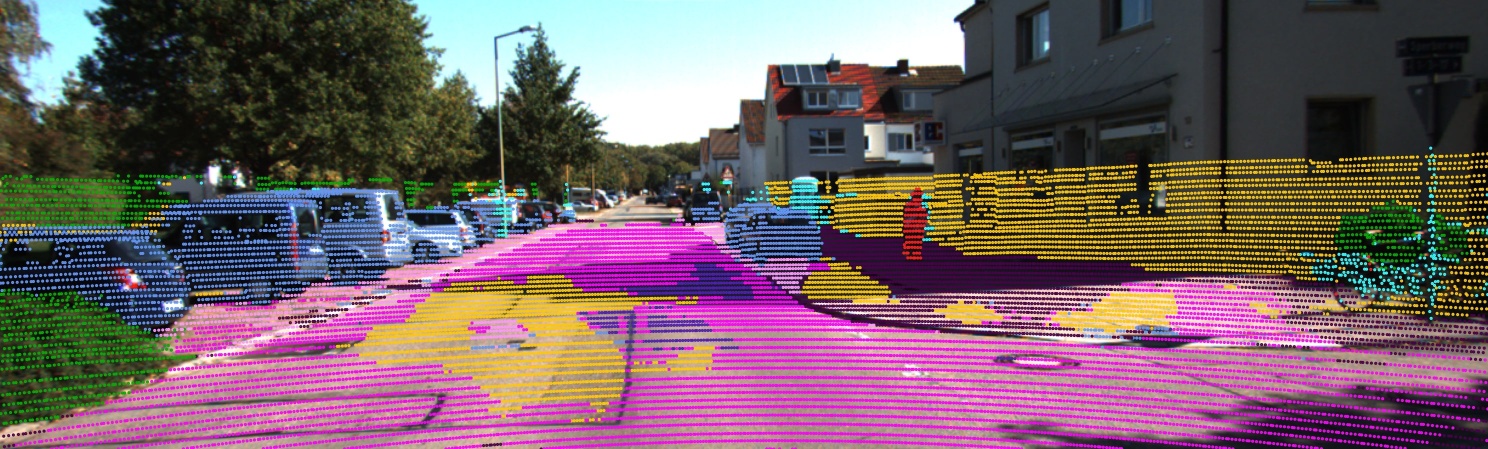}
        }
    }\\
    \vspace{-1em}
    \subfloat[(c) SUMMIT (Ours)]{
        \subfloat[2D]{
            \includegraphics[width=0.3\linewidth]{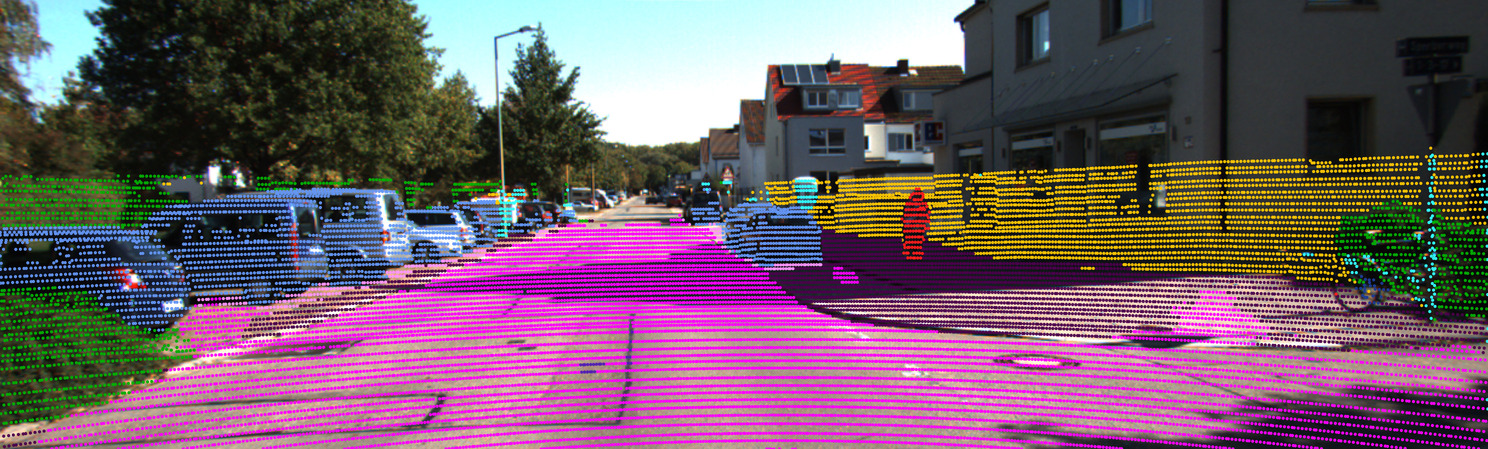}
        }
        \subfloat[3D]{
            \includegraphics[width=0.3\linewidth]{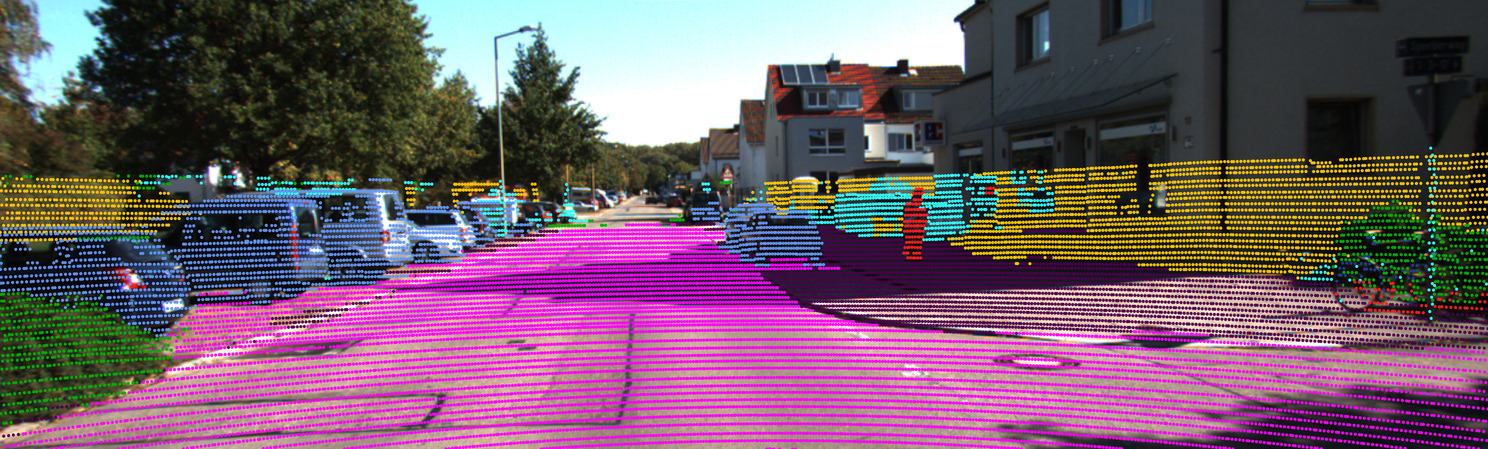}
        }
        \subfloat[2D+3D]{
            \includegraphics[width=0.3\linewidth]{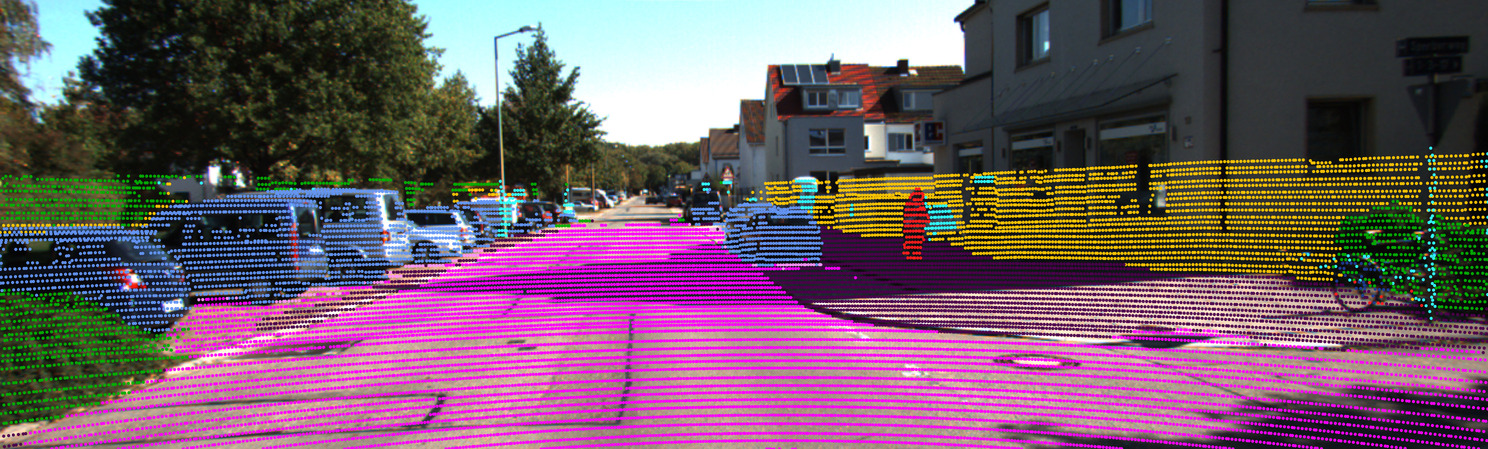}
        }
    }\\
    \caption{Qualitative results on the A2D2/SemanticKITTI experiment, comparing our method to the unadapted source model and xMUDA Source Free. Note the patches of misclassified road, which xMUDA SF exacerbates, but our method fixes. Images are best viewed digitally.}
    \vspace{-1em}
    \label{fig:a2d2_semantic_kitti_plot}
\end{figure}
\vskip 2pt
\noindent\textbf{Qualitative Results.}
We visualize the output of our adaptation methods as well as no adaptation and xMUDA SF baselines in Fig. \ref{fig:a2d2_semantic_kitti_plot}, for the adaptation experiment, and in Fig. \ref{fig:nuscenes_lidarseg_semantic_kitti_plot}, for the \emph{crossover} experiment. They can both be compared to the ground truth segmentation in Fig.~\ref{fig:semantic_kitti_ground_truth}. In both figures, we see that in the unadapted source, large portions of the road and sidewalk are misclassified in the 2D output. We see similar errors in the 3D and 2D+3D output. When the source-free xMUDA adaptation is applied, we see that the results can get worse, with even larger portions of the road and the sidewalk being misclassified. When we apply our method we see that most of the misclassifications are fixed, critically the road and sidewalk are now correctly classified. Some small portions of the building are misclassified, but they are relatively small regions. The combined output does fix some of the smaller misclassifications and the combination does not introduce any new errors. The images are best viewed digitally, where they can be zoomed in. A short video is provided in Apppendix \ref{apdx:videos}.

\section{Conclusion}
In this paper we introduced a new multi-modal UDA setting, Source-free Adaptation of Uni-modal Models to Multi-Modal Targets (SUMMIT), where source models are trained independently on each modality and source data are not needed during adaptation. We addressed this new setting by fusing information across modalities to improve pseudo-labeling. We proposed a data driven switching method that chooses between two complementary methods for fusing pseudo-labels across modalities, which provides an improvement of up to 12\% over competing baselines.

\noindent\textbf{Acknowledgments}
This work was supported in part by the US Department of Defense Laboratory University Collaboration Initiative program, National Science Foundation (Grant No. 172434 and 1901379), National Institute for Food and Agriculture (Award No. 2021-67022-33453) and the UC Multi-campus Research Programs Initiative.

{\small
\bibliographystyle{ieee_fullname}
\bibliography{references}
}

\appendix
\vspace{-4pt}
\section{Training Pseudo-Code}\label{apdx:pseudo_code}
We present the pseudo-code for our training algorithm in Algorithm \ref{summit_pseudocode}. The mathematical notation is the same as that described in Section 3.1 of the main paper. 
\vspace{-6pt}
\section{Additional training details}\label{apdx:hyper_parameters}
Most of our hyperparameters, shown in Table \ref{tab:hyperparameters}, we take directly from \cite{jaritz2020xmuda}, however we reduce the initial learning rate on the USA/Singapore and the Day/Night adaptation scenario. This is done because of the smaller domain gaps in these two adaptation scenarios. The initial learning rate in the case of the A2D2/SemanticKITTI adaptation scenario is relatively larger because of the larger domain gap. The value of $\lambda_{xM}$ is set according to \cite{jaritz2020xmuda}, including the lower value in the A2D2/SemanticKITTI adaptation scenario.
\begin{table}[H]
    \centering
    \small
    \caption{Here we show the hyper parameters for each experiment. The hyperparameters are the same for USA/Singapore \& the Day/Night experiments since we expect a similar domain gap. We modify the learning rate and $\lambda_{xM}$ slight for the A2D2/SemanticKITTI and all crossover experiments, since they have a larger domamin gap.}
    \resizebox{0.9\columnwidth}{!}{
    \begin{tabular}{lcccc}
      \toprule
       & USA/Singapore & Day/Night & A2D2/SemanticKITTI & Crossover\\
      \midrule
      Optimizer & Adam & Adam & Adam & Adam\\
      Learning Rate & 1e-5 & 1e-5 & 1e-3 & 1e-3\\
      $\beta_1$ & 0.9 & 0.9 & 0.9 & 0.9\\
      $\beta_2$ & 0.999 & 0.999 & 0.999 & 0.999\\
      Scheduler & MultiStep & MultiStep & MultiStep & MultiStep \\
      Learning Rate Decay & 0.1 & 0.1 & 0.1 & 0.1\\
      Milestone & 80K, 90K & 80K, 90K & 80K, 90K & 80K, 90K\\
      Max Iteration & 100K & 100K & 100K & 100K\\
      Batch Size & 8 & 8 & 8 & 8\\
      $\lambda_{xM}$ Target & 0.1 & 0.1 & 0.01 & 0.01\\
      \bottomrule
    \end{tabular}
    }
    \label{tab:hyperparameters}
\end{table}
\begin{algorithm}
	\caption{SUMMIT}
	\begin{algorithmic}[1]
	        \Require: Uni-Modal Source Models - $\mathcal{M}^{2D}$ \&  $\mathcal{M}^{3D}$, Source Model Metrics -  $Top1^{2D}$ \& $Top1^{3D}$,Multi-Modal Target Dataset - $\mathcal{D}_\mathcal{T}$
            \For {$\{x^{2D}_i, x^{3D}_i\} \in\mathcal{D}_{\mathcal{T}}$}
                \State $\tilde{y}^{2D}_i=\text{arg}\max_{k} \mathcal{M}^{2D}_k(x_i^{2D})$
		      \State $\tilde{y}^{2D}_i=\begin{cases}
                       \tilde{y}^{2D}_i, & \tilde{y}^{2D}_i, \geq \text{med}^{2D}_k\\
                       \textrm{ignore},& otherwise
                   \end{cases}$
  		    \State $\tilde{y}^{3D}_i=\text{arg}\max_{k} \mathcal{M}^{3D}_k(x_i^{3D})$
		      \State $\tilde{y}^{3D}_i=\begin{cases}
                       \tilde{y}^{3D}_i, & \tilde{y}^{3D}_i, \geq \text{med}^{3D}_k\\
                       \textrm{ignore},& otherwise
                   \end{cases}$
            \EndFor
            \State Source Agreement$=Top1^{2D}\cdot Top1^{3D}$
            \State Target Agreement$=\frac{\sum_{\tilde{y}^{2D}, \tilde{y}^{3D}} \mathbbm{1}(\tilde{y}^{2D} == \tilde{y}^{3D})}{|\mathcal{D}_\mathcal{T}|}$
            \If{$\frac{\text{Source Agreement}}{\text{Target Agreement}}\leq 0.5$}
                \For{$\tilde{y}^{2D}_i$ \& $\tilde{y}^{3D}_i$ }
                    \State $\tilde{y}_i=\begin{cases}
                    \tilde{y}_i^{2D},& \tilde{y}_i^{2D} = \tilde{y}_i^{3D}\\
                    \textrm{ignore},& \tilde{y}_i^{2D} \neq \tilde{y}_i^{3D}\\
                \end{cases}$
                \EndFor
            \Else
                \For {$\{x^{2D}_i, x^{3D}_i\} \in\mathcal{D}_{\mathcal{T}}$}
                    \State $w^{2D} = \frac{e^{-h(\mathcal{M}^{2D}(x^{2D}))}}{e^{-h(\mathcal{M}^{2D}(x^{2D}))} + e^{-h(\mathcal{M}^{3D}(x^{3D}))}}$
                    \State $w^{3D} = 1 - w^{2D}$
                    \State $\tilde{y}^{2D}_i=\mathcal{M}^{2D}(x^{2D})$
                    \State $\tilde{y}^{3D}_i=\mathcal{M}^{3D}(x^{3D})$
                    \State $p_i=w^{2D}\psi(\tilde{y}^{2D}_i)+w^{3D}\psi(\tilde{y}^{3D}_i)$
                    \State $\tilde{y}_i=\begin{cases}
                       \text{arg}\max_{k} p_i, & p_{i,k} \geq \text{med}^{p}_k\\
                       \textrm{ignore},& otherwise
                    \end{cases}$
                    \If {$\tilde{y}_i$ is ignored}
                        \State $R_{2D}=\frac{\mathcal{N}(f^{2D}(x^{2D}); \mu_{k_{2D}}^{2D}, \big(\sigma_{k_{2D}}^{2D}\big)^2)}{\mathcal{N}(f^{2D}(x^{2D}); \mu_{k_{3D}}^{2D}, \big(\sigma_{k_{3D}}^{2D}\big)^2)}$
                        \State $R_{3D}=\frac{\mathcal{N}(f^{3D}(x^{3D}); \mu_{k_{3D}}^{3D}, \big(\sigma_{k_{3D}}^{3D}\big)^2)}{\mathcal{N}(f^{3D}(x^{3D}); \mu_{k_{2D}}^{3D}, \big(\sigma_{k_{2D}}^{3D}\big)^2)}$
                        
                        \If {$R_{2D} \leq \tau$ and $R_{3D} > \tau$}
                            \State $\tilde{y}_i = \tilde{y}^{3D}_i$
                        \EndIf
                        \If{$R_{2D} > \tau$ and $R_{3D} \leq \tau$}
                            \State $\tilde{y}_i = \tilde{y}^{2D}_i$
                        \EndIf
                    \EndIf
                \EndFor
            \EndIf

            \For{$K$ iterations}
                \State Sample $\{{x^{2D}, x^{3D}, \tilde{y}}\}$ from $\mathcal{D}_\mathcal{T}$
                \State Calculate $\mathcal{L}_{tot}(\mathcal{M}^{2D}(x^{2D}_i), \mathcal{M}^{3D}(x^{3D}_i), \tilde{y})$
                \State Update $\mathcal{M}^{2D}, \mathcal{M}^{3D}$ to minimize $\mathcal{L}_{tot}$ 
            \EndFor
	\end{algorithmic}
	\label{summit_pseudocode}
\end{algorithm}

\begin{table*}[ht!]
    \centering
    \caption{In crossover experiments we see that the agreement filtering correctly labels a much larger portion of admitted samples. In contrast, the statistical fusion method admits many samples, but the pseudo-labels are generally incorrect. This indicates a larger portion of noise hurting the statistical fusion process.} 
    \resizebox{\textwidth}{!}{
    \begin{tabular}{lcccccccccccccc}
      \toprule
       & \multicolumn{3}{c}{\textbf{(USA-2D,Singapore-3D)/SemKITTI}} & \multicolumn{3}{c}{\textbf{(Singapore-2D,USA-3D)/SemKITTI}} & \multicolumn{3}{c}{\textbf{(Day-2D,Night-3D)/SemKITTI}} & 
       \multicolumn{3}{c}{\textbf{(Night-2D,Day-3D)/SemKITTI}}\\
      \cmidrule(lr){2-4}\cmidrule(lr){5-7}\cmidrule(lr){8-10}\cmidrule(lr){11-13}
      \textbf{Method} & Correct & Incorrect & Ignore & Correct & Incorrect & Ignore & Correct & Incorrect & Ignore & Correct & Incorrect & Ignore\\
      \midrule
      AF & 28.66 &  3.23 & 68.11 & 30.78 &  3.38 & 65.84 & 32.76 &  4.77 & 62.46 & 26.44 &  3.19 & 70.37\\
      EW &  7.49 & 85.28 &  7.23 &  4.26 & 89.92 &  5.82 &  4.41 & 89.97 &  5.62 &  1.41 & 94.88 &  3.71\\
      \bottomrule
    \end{tabular}\label{tab:crossover_filtering}
}
\end{table*}
\begin{table*}[ht!]
    \centering
    \caption{In crossover experiments we see that the source agreement is lower across the board. However, because of the domain gap between the source and target we see that the target agreement is far lower as well. This results in the crossover experiments consistently using the agreement filtering method.}
    \resizebox{\textwidth}{!}{
    \begin{tabular}{lcccc}
      \toprule
       &  \textbf{USA 2D, Sing. 3D} & \textbf{Sing. 2D, USA 3D} & \textbf{Day 2D, Night 3D} & \textbf{Night 2D, Day 3D}\\
      \midrule
      Source Agreement & 86.37 & 85.23 & 81.24 & 75.09\\
      Target Agreement & 38.60 & 30.60 & 32.80 & 35.00\\
      \midrule
      Ratio & 0.45 & 0.36 & 0.40 & 0.47\\
      \bottomrule
    \end{tabular}
    \label{tab:crossover_switching}
    }
\end{table*}

\section{Hypothesis Testing Threshold Analysis}\label{apdx:threshold}
In this section we present a sensitivity analysis of the hypothesis testing portion of entropy weighting to the threshold value. We show the results for thresholds of $0.5$, $1$, and $2$, which correspond to switching if the alternative is at least half as likely, just as likely, and twice as likely as the alternative. Unsurprisingly when $\tau=0.5$, we see the lowest performance, since in this case we chose the alternative when it's less likely than the null hypothesis. However, in this case the performance is still on par with the unadapted 2D and better than the unadapted 3D, which shows that even with a poorly tuned threshold the method does not hurt performance. In the case where $\tau=2$, we see that there are still minor improvements across the board. We see strongest improvement when $\tau=1$, which corresponds to only switching when one is more likely than the other.
\begin{table}[H]    
    \centering
    \caption{Here we present a analysis of the sensitivity of the hypothesis testing threshold. We note that while there is some variation in performance, our performance is still on par with unadapted performance. So while a better tuned value may help performance, poorly tuned values will not hurt performance.}
    \begin{tabular}{lccc}
      \toprule
       & \multicolumn{3}{c}{USA/Singapore}\\
      \midrule
      $\tau$ & 2D & 3D & 2D+3D\\
      \midrule
      0.5 & 49.25 & 50.18 & 53.39\\
        1 & 57.47 & 52.12 & 62.32\\
        2 & 50.27 & 47.97 & 56.19\\
      \bottomrule
    \end{tabular}\label{tab:sensativity}
\end{table}

\section{Crossover Automatic Switching Analysis}\label{apdx:switching}
We include here the full analysis of the pseudo-label switching method for the cross-over experiments. We present the full results in Table~\ref{tab:crossover_filtering}. We can see that the agreement filtering is indeed selected throughout the crossover experiments. We note the lower source agreement compared to the first three experiments. This is likely due to the Singapore \& Night splits of NuScenes having fewer samples to train the source model, about ~10K and ~3K respectively, compared to the USA \& Day splits which have ~16K and ~25K. However, since the target agreement is much lower the ratio between them still stays below our threshold of $0.5$, so AF is selected.

The accuracy of the filtered pseudo-labels is presented in Table~\ref{tab:crossover_filtering}. Once again we see that AF admits far fewer pseudo-labels, but the vast majority of these are correct. If we look at entropy weighting we see that in this case it has failed quite dramatically, admitting mostly incorrect labels. EW performs so poorly here because of the low source model performance, as can be seen in the no adaptation row of Table \ref{tab:crossover_ablation} in the main paper.

\section{Description of Attached Video}\label{apdx:videos}
We have uploaded two video examples of our method to youtube. Both videos show a clip from the SemanticKITTI dataset with the different colored points corresponding to the label predictions of a model adapted using agreement filtering. In \url{https://youtu.be/LDlLq9IdoAw} the source models are trained on the A2D2 dataset and in \url{https://youtu.be/ZFwWIXnKOEg} the 2D model is trained using the Day split of NuScenes and the 3D model is trained using the Night split, corresponding to the \emph{crossover} experiments. In both videos we show the predicted label for 2D \& 3D individually, the combined 2D+3D, and the ground truth classification. Please note that each frame is evaluated individually, making no use of any temporal correlations which we leave to future works.

\end{document}